\documentclass[12pt]{article}
\usepackage[utf8]{inputenc}
\usepackage{amsmath,comment}
\usepackage{graphicx} 
\usepackage{amssymb}
\usepackage{dsfont}
\usepackage{float}
\usepackage[legalpaper, margin=1in]{geometry} 
\usepackage{soul,color}
\usepackage{hyperref}
\usepackage{physics}
\usepackage{mathtools}
\usepackage[natbib=true,style=ieee]{biblatex}
\usepackage{lineno}
\usepackage[title]{appendix}
\usepackage{authblk}
\usepackage{enumitem}
\usepackage{amssymb}
\usepackage{pifont}
\usepackage{bbm}
\usepackage{trfsigns}
\usepackage{mathrsfs}

\usepackage{makecell} 
\date{November 7, 2023}

\DeclareRobustCommand{\rchi}{{\mathpalette\irchi\relax}}
\newcommand{\irchi}[2]{\raisebox{\depth}{$#1\chi$}} 

\newcommand{\xmark}{\ding{55}}%
\definecolor{cadmiumgreen}{rgb}{0.0, 0.42, 0.24} 

\title{Computing with Residue Numbers in \\ High-Dimensional Representation}

\author[1]{Christopher J. Kymn}
\author[2,3]{Denis Kleyko}
\author[4]{E. Paxon Frady}
\author[1]{Connor Bybee}
\author[1]{Pentti Kanerva}
\author[1,4]{Friedrich T. Sommer}
\author[1]{Bruno A. Olshausen}

\affil[1]{\footnotesize Redwood Center for Theoretical Neuroscience, University of California, Berkeley, CA}
\affil[2]{\footnotesize Centre for Applied Autonomous Sensor Systems, Örebro University, Sweden}
\affil[3]{\footnotesize Intelligent Systems Lab, Research Institutes of Sweden, Kista, Sweden}
\affil[4]{\footnotesize Neuromorphic Computing Lab, Intel, Santa Clara, CA}

\begin{document}
\maketitle

\begin{abstract}
We introduce \textit{Residue Hyperdimensional Computing}, a computing framework that unifies residue number systems with an algebra defined over random, high-dimensional vectors. We show how residue numbers can be represented as high-dimensional vectors in a manner that allows algebraic operations to be performed with component-wise, parallelizable operations on the vector elements. The resulting framework, when combined with an efficient method for factorizing high-dimensional vectors, can represent and operate on numerical values over a large dynamic range using vastly fewer resources than previous methods, and it exhibits impressive robustness to noise. We demonstrate the potential for this framework to solve computationally difficult problems in visual perception and combinatorial optimization, showing improvement over baseline methods. 
More broadly, the framework provides a possible account for the computational operations of grid cells in the brain, and it suggests new machine learning architectures for representing and manipulating numerical data.
\end{abstract}

\section{Introduction}

The problem of representing and computing on high-dimensional representations of numerical values---such as position, velocity, and color---is central to both machine learning and computational neuroscience. In machine learning, vector representations of numbers
are useful for defining position or function encodings in neural networks \citep{vaswani2017attention,sitzmann2020implicit,tancik2020fourier}, improving robustness to adversarial examples \citep{BuckmanThermometer2018}, and generating efficient classifiers  \citep{diao2021glvqhd}. In neuroscience, experimentalists seek to understand how populations of neurons in the brain represent and transform perceptual or cognitive variables, and so numerous theorists have constructed models for how these variables could be encoded in and decoded from high-dimensional vector encodings (e.g., \cite{pouget2000information,kriegeskorte2008representational,bordelon2022population}).

A particularly salient example of high-dimensional representation in neuroscience is the `grid cell' encoding of spatial position in the medial entorhinal cortex \citep{hafting2005microstructure}. Grid cells have multiple peaks in their firing rates that correlate with spatial positions arranged in a hexagonal lattice. While somewhat perplexing at first glance, the usefulness of such a coding scheme becomes apparent from how it functions as a population code. 
In comparison to a population of neurons with traditional unimodal encoding functions whose coding resolution increases linearly with the number of neurons, a grid cell population possesses a coding resolution that grows exponentially in the number of neurons \citep{mathis2012resolution}. In particular, Fiete and colleagues have emphasized that this advantage of grid cell encoding utilizes properties of {\em residue numbers} (see Section~\ref{sec:residue_def} below) \citep{fiete2008grid}.

Inspired by this observation, we propose a comprehensive algebraic framework for distributed neural computation based on residue number systems. Our novel framework builds on an existing algebraic framework for computing with high-dimensional random vectors. The idea was originally called computing in Holographic Reduced Representation \citep{PlateNested1994} and   is now referred to also as {\em Vector Symbolic Architectures (VSA)} \citep{GaylerJackendoff2003} and {\em Hyperdimensional Computing} \citep{kanerva2009hyperdimensional}. We call our new framework {\em Residue Hyperdimensional Computing} (RHC) and demonstrate that it inherits the computational advantages of both standard residue number systems and hyperdimensional computing. This enables fault-tolerant computations that can efficiently represent numbers and search over a large dynamic range with greatly reduced memory requirements. Furthermore, as we shall see, the new framework provides a useful formalism for understanding computations in grid cells.

To summarize, we list the four key coding properties we achieve with Residue Hyperdimensional Computing: (1) Algebraic structure: Simple operations on vectors perform addition and multiplication on encoded values, (2) Expressivity: Feasible encoding range scales better than linearly with dimension, (3) Efficient decoding: Required resources to decode scale better than linearly with encoding range, and (4) Robustness to noise. 
Although a number of previously proposed models achieve some of these properties (Table~\ref{table:compare}, Supplement~\ref{sec:otherlpe}), Residue Hyperdimensional Computing is the first, to our knowledge, to \textit{achieve all four of these desiderata}, as we shall now show. 

\section{Results}
\label{sec:results}

We first define the key concepts on which the Residue Hyperdimensional Computing framework is based (Section~\ref{sec:prelims}) and then describe the framework fully in Section \ref{sec:residue_def}. We then demonstrate its favorable encoding, decoding, and robustness properties (Section \ref{sec:scaling}), as well as how it can be extended to multiple dimensions (Section \ref{sec:multi-d}) and sub-integer encodings (Section \ref{sec:subint}). Of particular note, we construct a hexagonal residue encoding system, analogous to grid cell coordinates, that provides higher spatial resolution than square lattices. Finally, we describe how the framework can be applied to problems in visual scene analysis and combinatorial optimization (Section \ref{sec:applications}).

\subsection{Preliminary definitions}
\label{sec:prelims}

\textbf{\textit{Definition 2.1.1. }}A \textbf{Residue Number System} \citep{garner1959residue} encodes an integer $ x \in \mathbb{Z}$ by its value modulo $\{ m_1, m_2, \dots, m_K \}$, where the $m_k$ are the \textit{moduli} of the system. For example, relative to moduli $\{3,5,7\}$, $x=20$ would be encoded by the residue $[2,0,6]$---i.e., $[20\bmod3,\; 20\bmod5,\; 20\bmod7]$. The Chinese Remainder Theorem states that if the moduli are pairwise co-prime, then for any $x$ such that $0 \leq x < M \vcentcolon = \prod_k m_k$, the integer is uniquely encoded by its residue \citep{goldreich1999chinese}. From here on, we will assume that the pairwise co-prime condition is fulfilled. \\

\noindent \textbf{\textit{Definition 2.1.2.}} \textbf{Fractional Power Encoding (FPE)} \citep{PlateNested1994} defines a randomized mapping from an integer $x$ to a high-dimensional vector $\mathbf{z}(x)$.
Let $D$ be the dimension of the vector, with $D$ typically in the range $10^2 \leq D \leq 10^4$. The encoding scheme involves two steps. First, draw a random base vector, $\mathbf{z}$, defined as follows:
\begin{equation}
    \mathbf{z} = [e^{i \phi_{1}}, e^{i \phi_{2}}, \dots, e^{i \phi_{D}}]
\end{equation}
\noindent where each element $e^{i \phi_j}$ is a complex number with unit magnitude (a \textit{phasor}), and each $\phi_j$ is a random sample from a specified probability distribution. Second, define a function from $x$ to $\mathbb{C}^D$ via component-wise exponentiation of the random vector:
\begin{equation}
    \mathbf{z}(x) = \mathbf{z}^x = [e^{i \phi_{1}x}, e^{i \phi_{2}x}, \dots, e^{i \phi_{D}x}]
\end{equation}
\noindent \textbf{\textit{Definition 2.1.3.}} A \textbf{kernel}, $K(x_1,x_2)$, is a function $\rchi \cross \rchi \to \mathbb{R}$ that measures the similarity between two objects in a set $\rchi$ (e.g., vectors in $\mathbb{R}^n$). Notably, FPE implements kernel approximation \citep{plate2003holographic}, which is widely used in machine learning~\citep{rahimi2007random}. More specifically, we can induce a kernel based on the inner products of FPEs:
\begin{equation}
\label{eq:fpe}
K(x_1,x_2) = \frac{1}{D} \ \mathfrak{R} \{ \mathbf{z}(x_1)^T \ \overline{\mathbf{z}(x_2)} \}    
\end{equation}
\noindent where $\overline{\mathbf{z}(x_2)}$ is the complex conjugate of $\mathbf{z}(x_2)$. This defines a translation-invariant kernel $K(\Delta x)$ (where $\Delta x = x_1 - x_2$), which converges to a particular $K^*(\Delta x)$ as $D \to \infty$, where the shape of the kernel is determined by the probability distribution used to draw $\mathbf{z}$~\citep{frady2021computing, fradyfunctionsnice2022}.

\subsection{Residue Hyperdimensional Computing}
\label{sec:residue_def}
We now introduce how FPE can implement a residue number system. As a first step, we explain how FPE can implement \textit{congruence} (representing a remainder, modulo $m$). \\

\noindent \textbf{\textit{Definition 2.2.1.}} \textbf{Fractional Power Encoding, modulo $m$}: For $\mathbf{z}_m$, let the support of our probability distribution for the $\phi_j$ be the $m$-th roots of unity. In other words, each $\phi_j$ must be a multiple of $2\pi / m$.
Then congruent values are mapped to the same vector:
\begin{align*}
    \mathbf{z}_m(x+m) &=  [e^{i \phi_{1}(x+m)}, e^{i \phi_{2}(x+m)}, \dots, e^{i \phi_{D}(x+m)}] \\
    &= [e^{i\phi_{1}x}\cdot e^{i\phi_{1}m}, e^{i\phi_{2}x}\cdot e^{i\phi_{2}m}, \dots, e^{i\phi_{D}x}\cdot e^{i\phi_{D}m}] \\
    &= [e^{i \phi_{1}x}, e^{i \phi_{2}x}, \dots, e^{i \phi_{D}x}] \\
    &= \mathbf{z}_m(x)
\end{align*}
\noindent because $e^{i\phi_j m} = e^{2\pi \cdot i \cdot k}$ for some integer $k$, and $e^{2\pi \cdot i \cdot k} = 1$ for any integer $k$. Put another way, $\mathbf{z}_m(x)$ is a \textit{representation} (in the abstract algebraic sense) of the additive group of integers modulo $m$.

\begin{figure}[t]
\centering
\includegraphics[width=1.0\columnwidth]{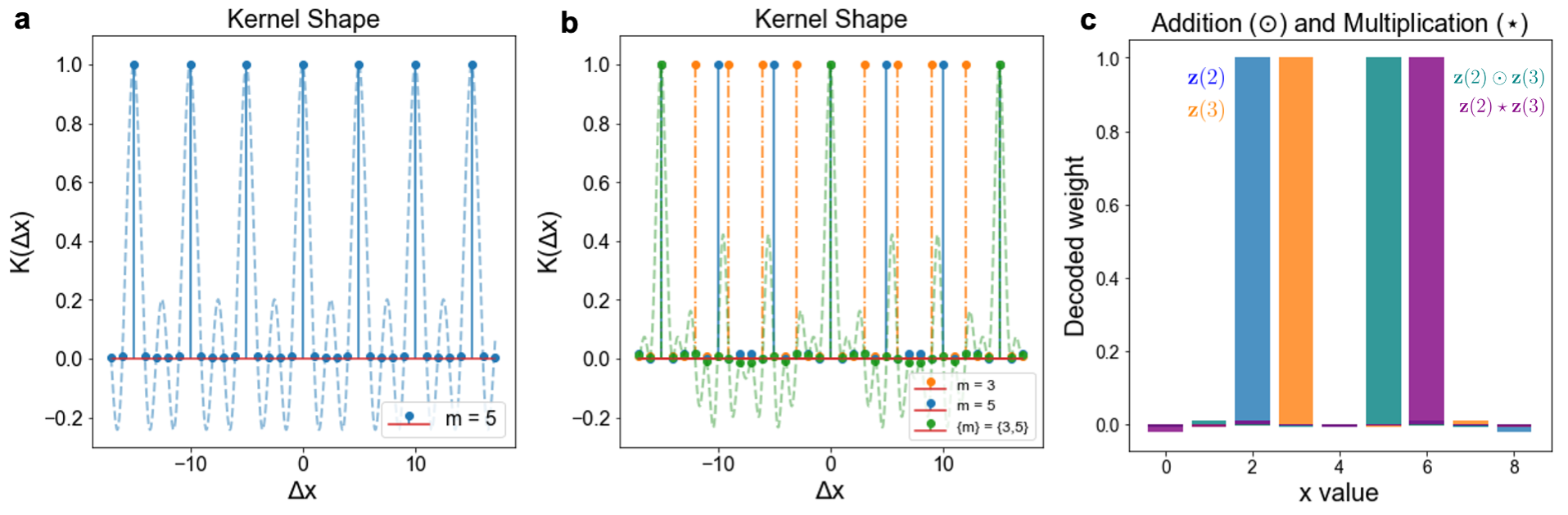}
\caption{\textbf{Residue Hyperdimensional Computing defines a kernel separating different remainder values, and it enables algebraic operations.} \textbf{a}, For fractional power encoding, modulo $m=5$, inner products between vectors reflect the similarity of points with the same remainder value and are quasi-orthogonal elsewhere. The light blue curve shows the kernel shape when $\Delta x$ is a real-valued scalar; integers occur approximately at zero crossings. A further derivation is provided in Supplement \ref{sec:periodic_kernel_math}. \textbf{b}, The kernel induced by an RHC vector (green) is the product kernel of the moduli used to form it (orange, blue). \textbf{c}, Demonstration of addition and multiplication. Blue and orange show encodings of 2 and 3, respectively. Teal shows the decoded value of 2+3 (i.e., 5); purple shows decoded value of $2\times3$ (i.e., 6).}
\label{fig:kernelalgebra}
\end{figure}

The kernel induced by $\mathbf{z}_m(\Delta x)$ is $1$ if $\Delta x = 0$ (mod $m$), and $\approx 0$ otherwise, as shown in Figure~\ref{fig:kernelalgebra}a. This is highly useful, because it implies that distinct integers behave as quasi-orthogonal vectors, just like symbols in hyperdimensional computing. Unlike symbols, however, we can perform algebraic manipulations transforming one integer into another. \\

\noindent \textbf{\textit{Definition 2.2.2.}} \textbf{Residue Hyperdimensional Computing}: Let $\mathbf{z}_{m_1}$, $\mathbf{z}_{m_2}$, $\dots$, $\mathbf{z}_{m_K}$ denote FPE vectors with moduli $m_1, m_2, \dots, m_K$ respectively. Let $\odot$ denote component-wise multiplication (a.k.a. Hadamard product). Then, we encode an integer $x$ by combining our modulo representations via the Hadamard product:
\begin{equation}
    \mathbf{z}(x) = \bigodot_{k=1}^K \mathbf{z}_{m_k}(x) 
\label{eq:residuehdc}
\end{equation}

The above encoding represents the remainder of $x$, because each $\mathbf{z}_{m_k}$ represents its value modulo $m_k$. The code is fully distributed, as every element of the vector $\mathbf{z}$ contains information about each encoding vector $\mathbf{z}_{m_k}(x)$. By contrast, typical implementations of residue number systems compartmentalize information about individual moduli \citep{omondi2007residue}.

The kernel induced by $\mathbf{z}(\Delta x)$ is $1$ if $\Delta x = 0$ (mod $\prod_k m_k = M$), and $\approx 0$ for other integer intervals $\Delta x$ (Figure~\ref{fig:kernelalgebra}b). This means that the kernel maps different remainders of our residue number system to quasi-orthogonal directions in high-dimensional vector space, and enables computing in superposition over these variables. Examples of possible applications enabled by such a scheme are presented in Section \ref{sec:applications}.

The hallmark of a residue number system is \textit{carry-free arithmetic}; that is, addition, subtraction, and multiplication can be performed component-wise on the remainders. This enables residue number systems to be highly parallel, avoiding carryover operations required in binary number systems. \textit{Residue Hyperdimensional Computing implements arithmetic with component-wise operations, thus inheriting the premier computational property of residue number systems.}

Addition is defined as the Hadamard product between vectors, that is $\mathbf{z}(x_1 + x_2) = \mathbf{z}(x_1) \odot \mathbf{z}(x_2)$ (Methods \ref{sec:additivebinding}). This follows from the fact that component-wise multiplication of phasors corresponds to phase addition, and that component-wise multiplication is commutative. Subtraction is defined by addition of the additive inverse.

Next, we define a second binding operation that implements multiplication, denoted as $\star$: $\mathbf{z}(x_1 \cdot x_2) = \mathbf{z}(x_1) \star \mathbf{z}(x_2)$. Just as variable addition is implemented by element-wise multiplication, variable multiplication is implemented by another element-wise operation, this one involving exponentiation (Methods \ref{sec:secondbinding}). Here, it is crucial that our encoding function is restricted to integers as its domain, because multiplication is commutative and integer powers commute (that is, $(c^{x_1})^{x_2} = c^{x_1 x_2} = (c^{x_2})^{x_1}$ for $c \in \mathbb{C}$ and integers $x_1,x_2$). We show how this definition for integer multiplication can be generalized to multiplication for vector encodings $\mathbf{z}(x_1)$ and $\mathbf{z}(x_2)$, without invoking the costly step of first decoding the integers back from the vectors.

Division is not well-defined for residue number systems, because integers are not closed under division. Still, when a modulus $m_k$ is prime, multiplications by non-zero integers are invertible, because each non-zero integer has a unique multiplicative inverse with respect to $m_k$.

\textit{The existence of two distinct binding operators for addition and multiplication is a new contribution to hyperdimensional computing.} Previous formulations only supported addition, or only multiplication via addition after taking logarithmic transformations \citep{kleyko2022vector}. Consequently, we can now formulate a fully distributed residue number system that inherits the benefits of computing with high-dimensional vectors. 

\subsection{The resonator network enables efficient decoding of residue \\ numbers}
\label{sec:scaling}

Given the vector representation of a residue number, $\mathbf{z}(x)$, how do we decode it to recover its value, $x$? One method for decoding commonly used in hyperdimensional computing is \textit{codebook decoding}~\citep{kleyko2023efficient}, which involves taking the inner product of $\mathbf{z}(x)$ with a set of $M$ codebook vectors with known values of $x$.  However, this procedure requires $\mathcal{O}(M*D)$ storage and $M$ inner product evaluations.

\begin{figure}[t]
\centering
\includegraphics[width=1.0\columnwidth]{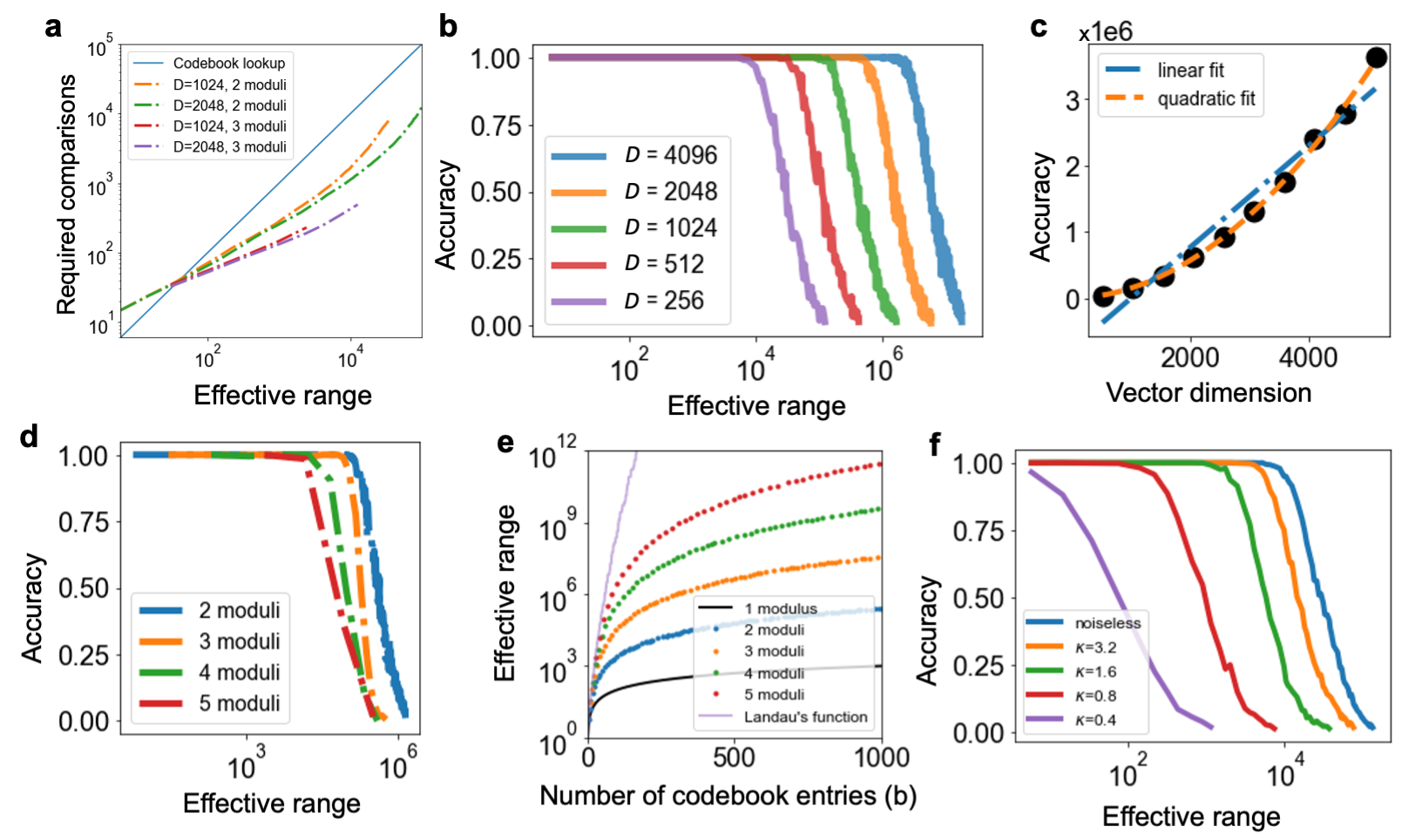}
\caption{\textbf{The resonator network can efficiently recover moduli of different numbers.} \textbf{a}, The resonator network outperforms codebook decoding by over an order of magnitude when the effective range is within resonator network capacity. \textbf{b}, Demonstration of resonator network capacity. For a fixed vector dimension, accuracy remains high up to a given range, before gradually falling off. \textbf{c}, Scaling of resonator network capacity, $C$, as a function of dimension, $D$, is well-described by a quadratic fit (orange dashed line; cf. linear fit in blue dash-dotted line). Quadratic fit coefficients are $(1.3 \times 10^{-1})D^{2} + (2.4 \times 10^1)D -3.7 \times 10^3$, linear fit coefficients are $(7.6 \times 10^{2})D - 7.6 \times 10^{5}.$ \textbf{d}, Resonator network performance is slightly worse for a higher number of moduli, $K$, but \textbf{e}, an advantage of lower moduli is a much higher effective range given encoding resources. \textbf{f}, The capacity of the resonator network remains high even in the presence of large amounts of phase noise.
}
\label{fig:capacity}
\end{figure}

Fortunately, we can improve the situation by utilizing the fact that residue numbers break the dynamic range of a number into a set of smaller numbers, each with lower overall dynamic range.  For example, a number with dynamic range of 105, when represented modulo $[3, \ 5, \ 7]$, consumes a total dynamic range of $3+5+7=15$.  This in turn reduces both the memory and computation resources for decoding by a factor of $105/15=7$.  To make this work though requires that we invert Equation~\ref{eq:residuehdc}---that is, we must factorize $\mathbf{z}(x)$ into the set of constituent vectors $\{ \mathbf{z}_{m_1} (x),\mathbf{z}_{m_2}(x),\dots,\mathbf{z}_{m_K}(x) \}$ representing $x$ modulo $m_k$, from which $x$ can be easily recovered. For this we can use a {\em resonator network} \citep{FradyResonator2020,kent2020resonator}, a recently discovered method for efficiently factorizing vectors in hyperdimensional computing.
Figure~\ref{fig:capacity}a shows that for a range of $M$ values, the resonator network can recover vectors over an order of magnitude faster than standard codebook decoding. Two parameters that contribute to this are the vector dimension ($D$) and number of moduli ($K$).

To evaluate the dependence of resonator decoding on vector dimension, we fix the number of moduli ($K=2$) and calculate the empirical accuracy (Figure~\ref{fig:capacity}b) of the resonator on the effective range $M$. We find that for a fixed $D$, the accuracy remains almost perfect up to a certain range of $M$, after which accuracy rapidly decays. To evaluate scaling with $D$, we define the capacity $C$ of a dimension to be the largest $M$ up to which empirical accuracy is at least 95 percent. We find that the scaling of $C(D)$ is well-fit with a quadratic polynomial (Figure~\ref{fig:capacity}c), consistent with previous scaling laws studied for a resonator network with two states per component \citep{kent2020resonator}. Further tests with higher dimensions would help confirm quadratic scaling, but even the linear scaling has high slope (and note that $C(4096) > 2 \times 10^6$).

To evaluate dependence on the number of moduli, we fix $D=1024$ and vary $K$. We find that resonator capacity decreases as $K$ increases (Figure~\ref{fig:capacity}d), also consistent with prior work \citep{kent2020resonator}. Still, we emphasize that resonators with higher $K$ have two advantages: decreased computation per decoding (Figure~\ref{fig:capacity}a), and decreased memory requirements. The resonator requires only $\sum_k m_k = b$ codebook vectors, rather than $\prod_k m_k = M$. This means that increasing $K$ can increase the effective range by \textit{several orders of magnitude} given a fixed codebook budget (Figure~\ref{fig:capacity}e). Remarkably, the maximal $M$ for a given $b$ is given by Landau's function $g(b)$, which scales as $g(b) = e^{(1+o(1))\sqrt{b\ln b}}$ \citep{landau1903maximalordnung}. This implies an exponential scaling between storage required and effective range, if we proffer sufficient $K$ to achieve it.

Finally, we evaluate how robust resonator network decoding is to noise. We draw phase noise from a von Mises distribution with mean 0 and concentration $\kappa$; higher $\kappa$ indicates less noise. In Figure~\ref{fig:capacity}f, we observe that performance degrades gradually as a function of noise, yet capacity remains remarkably high even at high noise levels.

\subsection{Generalization to multiple dimensions}
\label{sec:multi-d}

\subsubsection{Cartesian representations of $\mathbf{Z}^n$}
\label{sec:cartesian-coord}

Next, we generalize Residue Hyperdimensional Computing from scalars to multi-dimensional variables, showing that the core operations and principles still apply. Let $\mathbf{x} \in \mathbb{Z}^n$ be a low-dimensional vector.  Let $x_1, x_2, \dots, x_n$ denote the components of $\mathbf{x}$. To encode $\mathbf{x}$ with a single high-dimensional vector $\mathbf{z} \in \mathbb{V}^D$ ($D >> n)$ we form the encoding $\mathbf{z}(x)$ by taking the Hadamard product of the distributed representations of individual components:
\begin{equation}
    \mathbf{z}(\mathbf{x}) = \mathbf{z}_1(x_1) \odot \mathbf{z}_2(x_2) \odot \dots \odot \mathbf{z}_n(x_n)
    \label{eq:znfpe}
\end{equation}

\noindent where each $\mathbf{z}_i$ is a random vector as generated for the residue representation of a single number in Equation~(\ref{eq:residuehdc}). (Each $\mathbf{z}_i$ is created from binding multiple vectors for each of the moduli, distinct for each dimension $i$.) Since the binding operation is commutative, we can rearrange our terms to show the following useful properties hold:
\begin{enumerate}[label=(\alph*)]
    \item The Hadamard product ($\odot$) performs vector addition: $\mathbf{z}(\mathbf{x}) \odot \mathbf{z}(\mathbf{x'}) = \mathbf{z}(\mathbf{x + x'})$.
    \item The multiplicative binding operation ($\star$) performs component-wise multiplication of $\mathbf{x}$ and $\mathbf{x'}$: $\mathbf{z}(\mathbf{x}) \star \mathbf{z}(\mathbf{x'}) = \mathbf{z}(\mathbf{x} \odot \mathbf{x'})$.
    \item The kernel induced by $\mathbf{z}$ is the product of the kernels of the individual components: $K_z(\mathbf{\Delta x}) = \prod_{i=1}^{n} K_{z_i}(\Delta x_i)$, where $\mathbf{\Delta x} = \mathbf{x} - \mathbf{x'}$.
\end{enumerate}

\noindent While (a) and (c) are general properties of FPE, (b) is once again unique to Residue Hyperdimensional Computing. In addition, the savings in decoding resources and computation required (Figure~\ref{fig:capacity}) also scale in higher dimensions, as utilized in Section~\ref{sec:applications}. 

\subsubsection{Hexagonal coordinate systems}
\label{sec:hexcoord}
When working in a multi-dimensional space, there are multiple alternatives to a Cartesian coordinate system.  For example, grid cells in medial entorhinal cortex encode spatial location with a hexagonal coordinate system; research in theoretical neuroscience suggests that this is because such a tiling of space has the highest resolution (Fisher information) in 2D space \citep{mathis2015probable}. As an illustrative example, we show that Residue Hyperdimensional Computing can also implement hexagonal coordinate systems, and that such a hexagonal lattice retains coding advantages over square lattices. Of particular note, we formulate a self-consistent encoding that extends a residue number system to non-negative hexagonal coordinates.

\begin{figure}[t]
\centering
\includegraphics[width=1.0\columnwidth]{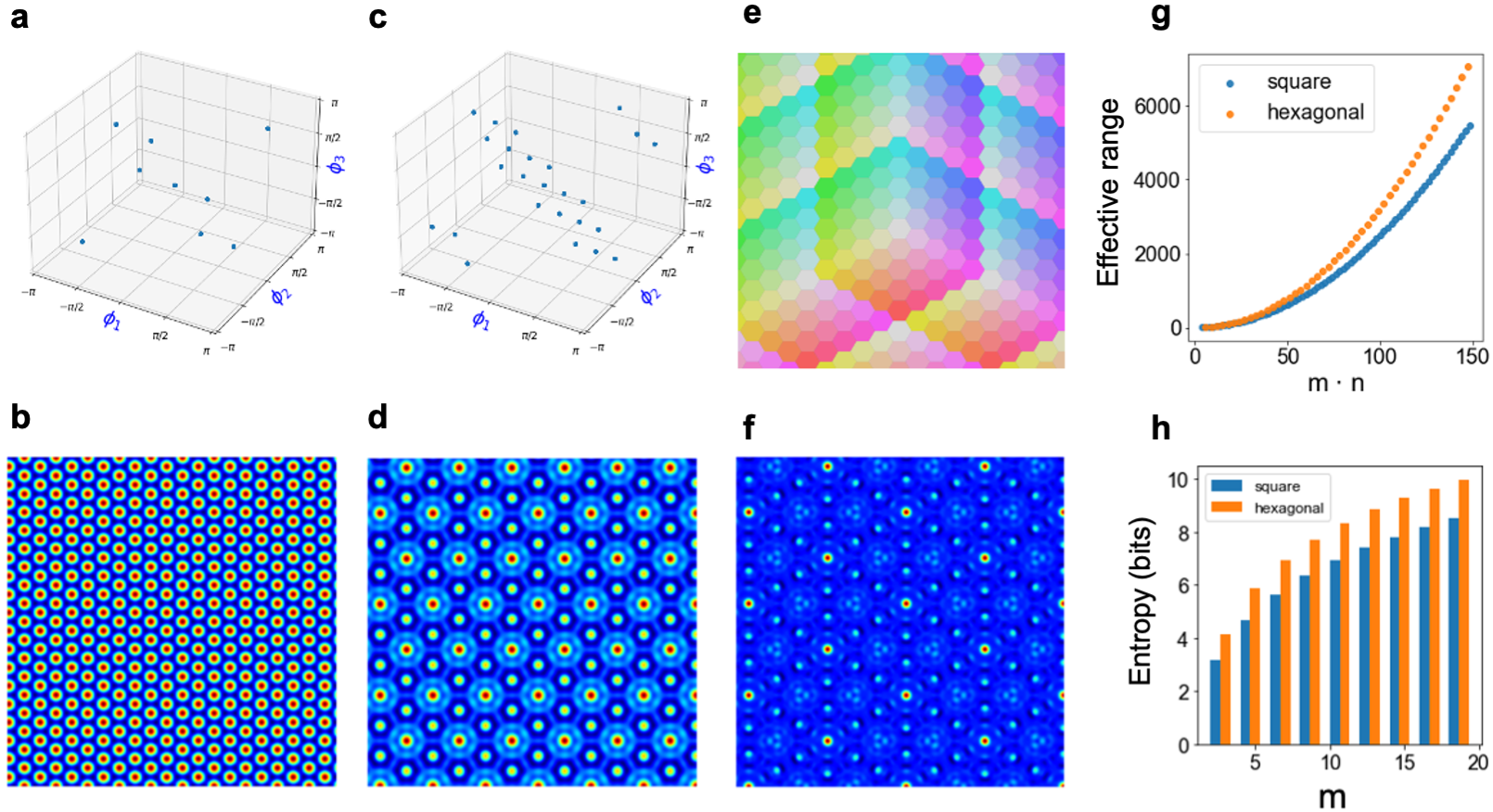}
\caption{\textbf{Definition of a residue number system in non-negative hexagonal coordinates.} \textbf{a} and \textbf{c} show discrete phase distributions chosen for a 3D coordinate system of a 2D space with moduli 3 and 5, respectively. In addition to requiring that phases are drawn from the $m$-th roots of unity, we enforce that the three phases sum to $0 \ (\text{mod} \ 2\pi)$. \textbf{b} and \textbf{d} show the respective kernels generated by these phase distributions. \textbf{e}, An example of the Voronoi tesselation of different states composed of a hexagonal coordinate system with modulus 5. Each color corresponds to a different state representation in the vector space of integers. \textbf{f}, The kernel induced by a hexagonal residue HD vector (period of $3 \cdot 5 = 15$). \textbf{g}, Compared to square encodings of space, hexagonal encodings approximately triple the effective range of encodable states with only a 50 percent increase in required storage space. \textbf{h}, The Shannon entropy of the hexagonal code is higher than that of the square code with the same modulus, reflecting the benefit of hexagonal packing.}
\label{fig:2d}
\end{figure}

To encode a two-dimensional position into a three-coordinate frame requires two steps. First, we project the 2D vector $\mathbf{x}$ into a 3D vector $\mathbf{y}$ with unit vectors whose angles each differ by $\frac{2\pi}{3}$ (Methods~\ref{sec:hexeq}). This coordinate representation is known as the `Mercedes-Benz' frame in $\mathbb{R}^2$; it is well-studied in signal processing \citep{malozemov2009equiangular} and has attracted recent interest in hyperdimensional computing (e.g., \citep{komer2020biologically,frady2021computing}). For our purposes, this step is necessary, but not sufficient, because projections can result in negative values. Second, to rectify values, we encode $\mathbf{y}$ with the method described in Section \ref{sec:cartesian-coord}, but with the additional constraint that $\mathbf{z}([1,1,1]) = \mathbf{z}([0,0,0])$, ensuring that every state with negative coordinates has an equivalent representation to one with non-negative coordinates (Figures~\ref{fig:2d}a and~\ref{fig:2d}c, Methods \ref{sec:hexeq}). This constraint also reflects the fact that equal movement in every direction cancels out, and thus it enforces that different paths to the same 2D position result in the same high-dimensional encoding. The kernels induced by vectors of individual moduli (Figures~\ref{fig:2d}b and~\ref{fig:2d}d) and by the residue vector (Figure~\ref{fig:2d}f) exhibit the six-fold symmetry characteristic of hexagonal lattices and grid cells \citep{hafting2005microstructure}.

We can, therefore, represent the hexagonal coordinate system with a Voronoi tesselation (Figure~\ref{fig:2d}e) in which different regions of space are mapped to their nearest integer-valued 3D coordinate. A hexagonal system with modulus $m$ has $3m^2-3m+1$ distinct states and requires $3m$ codebook vectors, whereas a square lattice has $m^2$ distinct states and requires $2m$ codebook vectors (Figure~\ref{fig:2d}g). Thus, the hexagonal system achieves better spatial resolution (it is a higher entropy code with regards to space) than a square lattice (Figure~\ref{fig:2d}h) for the same number of resources.

\subsection{Extensions to sub-integer decoding resolution}
\label{sec:subint}

\begin{figure}[t]
\centering
\includegraphics[width=1.0\columnwidth]{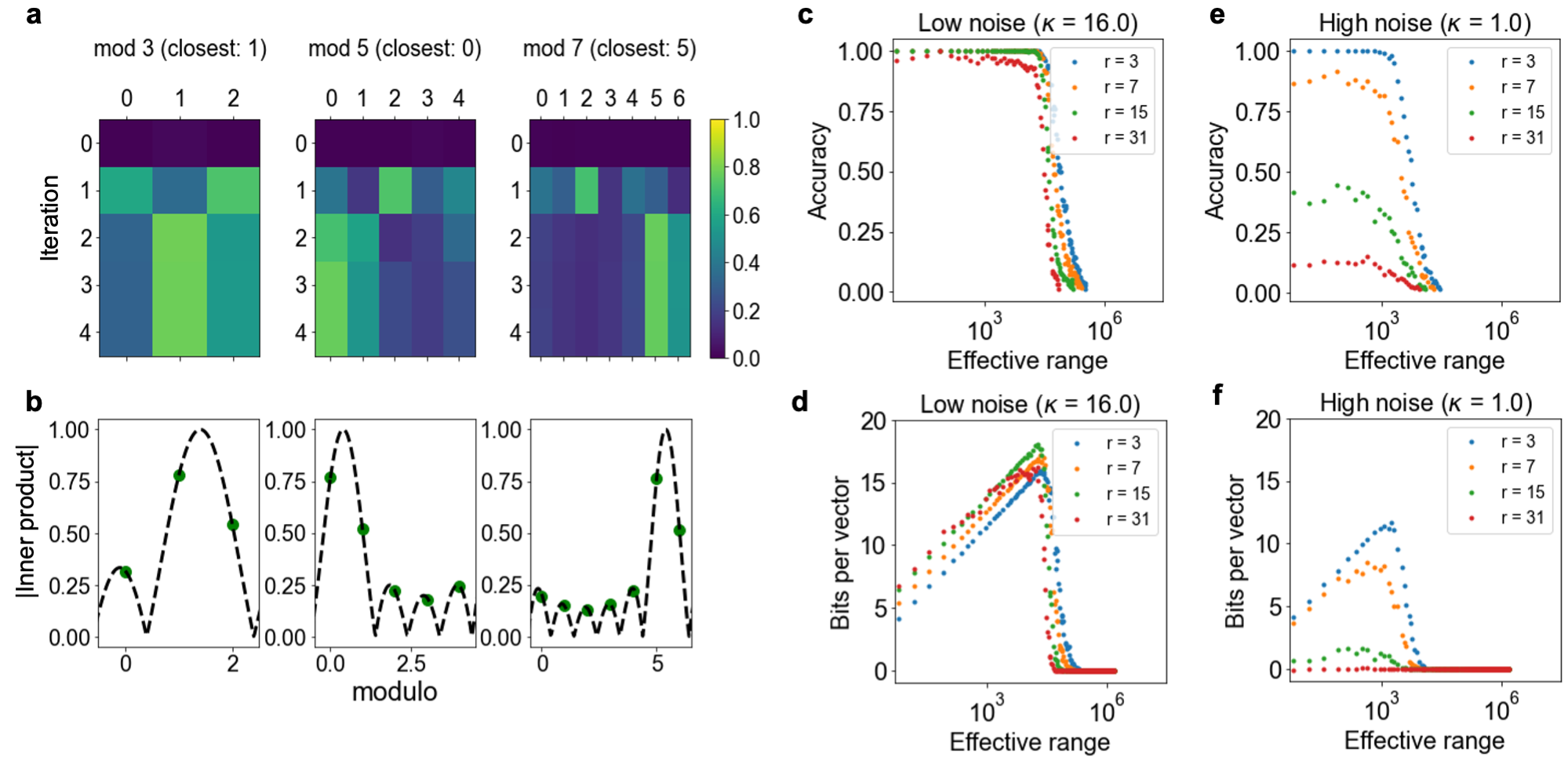}
\caption{\textbf{The resonator network enables retrieval of fractional (sub-integer) values.} \textbf{a}, Example of the resonator network converging for $\mathbf{z}(x)$, $x = 40.4$. \textbf{b}, Inner product values decoded by the resonator network are predicted by fractional offsets from a Dirac comb convolved with a sinc function. \textbf{c}, Sub-integer encoding accuracy under a low noise regime for a RHC vector ($r$ denotes number of sub-integer partitions). \textbf{d}, Bits per vector for different fractional decodings. \textbf{e, f}, same as \textbf{c} and \textbf{d}, respectively, but under higher noise conditions.}
\label{fig:fractional}
\end{figure}

In previous sections, we worked exclusively with integer states and residue number systems implementing them. Intriguingly, however, we can extend our definition of FPE to rational numbers (Methods \ref{sec:rationalfpe}), and the resonator converges to FPE encodings of non-integers, even when codebooks contain only encodings of integers (Figures~\ref{fig:fractional}a and b). Strictly speaking, such extensions beyond integers are not residue number systems, and multiplicative binding is no longer well-defined. However, extensions towards sub-integer resolution have been considered in theoretical analyses of grid cells, e.g., in \citep{fiete2008grid,sreenivasan2011grid}, and we show that the resonator dynamics achieve this sub-integer resolution.

An efficient procedure for decoding with sub-integer precision is suggested by Figure~\ref{fig:fractional}b. The inner products between codebook states and the final resonator network state are well-described by evaluations of a Dirac comb convolved with a sinc function (Supplement \ref{sec:periodic_kernel_math}). For integer encodings, this function would evaluate to 0 for all features not near the peak, but for non-integers this is no longer the case. Still, we can find the sub-integer offset that best matches the resonator network state in order to decode the sub-integer value (Methods \ref{sec:subintdecoding}). 

Phase noise is the limiting factor in decoding sub-integer states. To quantify this more rigorously, we evaluate a resonator network with varying effective ranges $M$ under different noise regimes ($\kappa = 16$ and $1$, respectively). We then split each unit interval into $r$ partitions, so that there are $M \cdot r$ distinct numbers represented.
Figures~\ref{fig:fractional}c and~\ref{fig:fractional}e show the accuracy of decoding for a different number of partitions with $\kappa = 16$ and $\kappa = 1$, respectively. To account for accuracy and the number of different states distinguished, we also report the `bits per vector' metric \citep{frady2018theory} in Figures~\ref{fig:fractional}d and~\ref{fig:fractional}f. This metric validates that with lower noise, we can more reliably decode a higher number of states.

\subsection{Applications}
\label{sec:applications}

\subsubsection{Efficient disentangling of object shape and pose from images}
\label{sec:vis}
\begin{figure}[t]
\centering
\includegraphics[width=1.0\columnwidth]{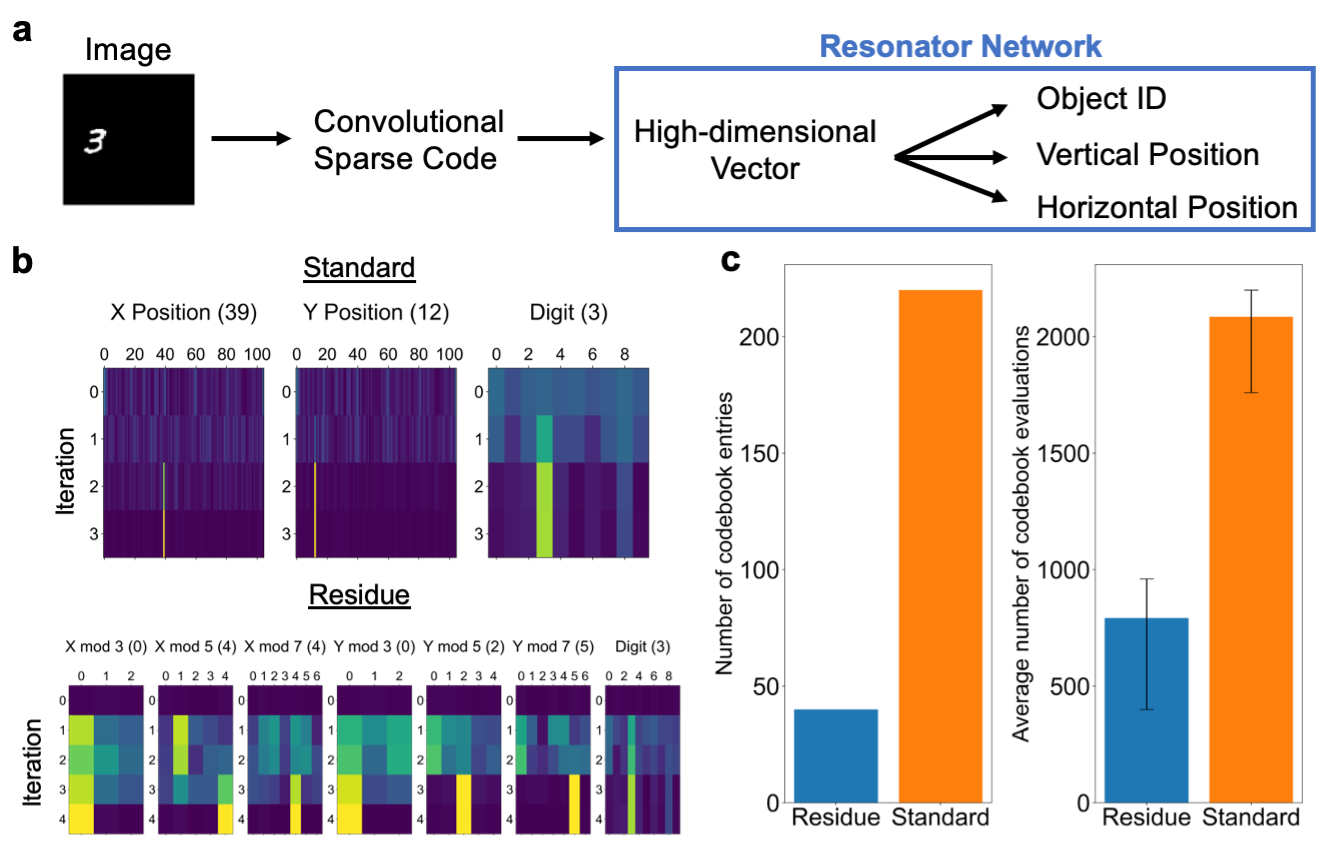}
\caption{\textbf{Residue Hyperdimensional Computing enables efficient disentangling of images.} \textbf{a}, Processing pipeline: an image is first represented in terms of its local shape features via convolutional sparse coding, then converted to a high-dimensional vector $\mathbf{z}$ by superimposing the residue number encodings of the positions of each feature in the image, which is finally factorized into object identity and position via a resonator network. \textbf{b}, Simulations of the resonator network on visual scenes without the residue number encoding (Standard) and with the residue number encoding (Residue). \textbf{c}, With a residue number system, the resonator network requires less memory overhead (40 vs.\ 220 codebook vectors) and less total computation to converge to the correct solution (792.4 vs.\ 2085.6 average codebook evaluations). Error bars show the 25\textsuperscript{th} and 75\textsuperscript{th} percentile of the number of evaluations.}
\label{fig:visual}
\end{figure}

Here we study the disentangling problem in vision---that is, the task of recovering the underlying components of a scene given only the image pixel values. Such problems abound in visual perception; examples include inferring structure-from-motion or separating spectral reflectance properties from shading due to 3D shape and lighting. How brains disentangle these factors of variation in general is unknown, and it is computationally challenging due to the combinatorial explosion in how factors combine to create any given scene \citep{olshausen2014}.

Here we demonstrate how Residue Hyperdimensional Computing can efficiently tackle the combinatorial complexity inherent in visual perception by considering a simple case of images containing three factors of variation: object shape, horizontal position, and vertical position. Let $O, H, V$ be finite sets listing the possible vectors for each factor. The goal is to infer the $O_a \in O, H_b \in H, V_c \in V$ provided an image, $I$. In this setup, the search space is $|O|\cdot|H|\cdot|V|$, and in our example (Figure~\ref{fig:visual}), $|O| = 10$, and $|H| = |V| = 105$, giving a search space of $\sim 10^5$. 

We solve this problem in two stages. First, we form a latent, feature-based representation of the image via convolutional sparse coding (Methods \ref{sec:csc}). This step mirrors the neural representation in primary visual cortex, which is hypothesized to describe image content in terms of a small number of image features~\citep{olshausen1996emergence}.  We observe that this step is useful as it helps to decorrelate image patterns, thus achieving higher accuracy and faster convergence for the resonator network. 
 
Second, we encode the latent feature representation into a high-dimensional vector that can be subsequently factorized into its components ($O_a$, $H_b$, $V_c$) via a resonator network. This is accomplished, following \citep{renner2022neuromorphic}, by superimposing the residue number encodings of the position of each image feature into a single scene vector, $\mathbf{s}$ (Methods~\ref{sec:csc}, equation~\ref{eq:image-encoding}).  The resulting vector, $\mathbf{s}$, can be expressed equivalently as a product of vectors representing object shape and position, and thus the problem of disentangling these factors essentially amounts to a vector factorization problem.

The standard way to factorize the scene vector
(e.g., as in \cite{renner2022neuromorphic}) would be to use three codebooks corresponding to shape, horizontal position and vertical position, for a total of $10+105*2 = 220$ codebook vectors. By contrast, a residue number system with moduli $\{3,5,7\}$ uses 7 factors but only $10+(3+5+7)*2 = 40$ vectors. Example runs of both problem setups are shown in Figure~\ref{fig:visual}b.

\textit{Figure~\ref{fig:visual}c demonstrates the two main advantages of the residue resonator compared to the standard resonator baseline: a reduction in both memory requirements (as just described) and the required number of iterations.} Whereas the standard resonator takes over $\approx 2{,}000$ codebook evaluations on average in our simulations, the residue resonator averages only $\approx 800$ codebook evaluations. (Both dramatically improve over the brute force search, which requires $110{,}250$ codebook evaluations). The key lesson is that this simple change to a resonator network leads to a multiplicative decrease in the number of computations required.

\subsubsection{Generating exact solutions to the subset sum problem}
\label{sec:subsetsum}

\begin{figure}[t]
\centering
\includegraphics[width=1.0\columnwidth]{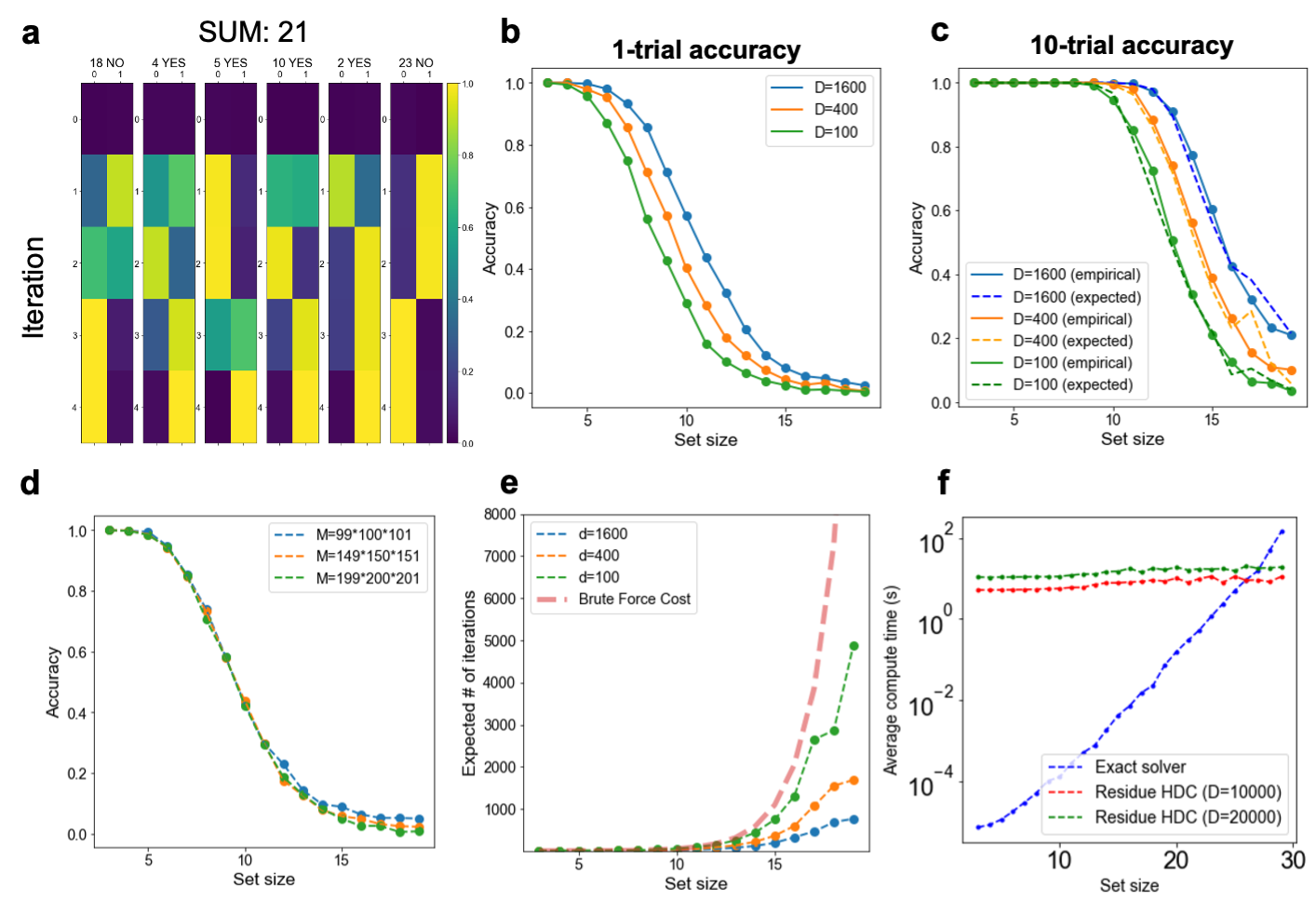}
\caption{\textbf{The resonator network, with Residue Hyperdimensional Computing, enables successful searches for solutions to the subset sum problem.} \textbf{a}, Demonstration of the resonator network on the subset sum problem, with $S = \{18, 4, 5, 10, 2, 23\}$. The resonator network converges to the correct solution after a few iterations. \textbf{b}, Performance of the resonator network on randomly selected subset sum problems with a fixed set size. \textbf{c}, The probability of success scales as expected with independent trials on each start. \textbf{d}, Success of the resonator network on the subset sum problem depends on the range of items indexed in the subset sum problem (larger ranges are harder to factorize). \textbf{e}, Performance of the resonator network in terms of the expected number of iterations compared to chance. The expected number of iterations for the resonator network scales favorably compared to brute force search and improves with higher encoding dimension. \textbf{f}, Comparison of average compute time of the resonator network versus an exact algorithm. The initialization cost of setting up the resonator network is higher; even so, for large set sizes, the resonator network is faster.}
\label{fig:subset}
\end{figure}

Here we apply Residue Hyperdimensional Computing to the subset sum problem. Formally, the problem asks if a multiset of integers, $S$, contains a subset $S^*$ that sums to a target integer $T$. A further demand is to return $S^*$ if it exists. When all integers are positive, the subset-sum problem is NP-complete \citep{kleinberg2006algorithm}. It is a useful case to consider because there are well-known polynomial-time reductions from other NP-complete problems (e.g., 3-SAT) to subset sum \citep{karp1972chapter}.

To find solutions to the subset sum problem, we encode $T$ as a vector, $\mathbf{z}(T)$, and use $|S|$ factors. Each factor, $F_k$, has a codebook of two items: an identity vector $\mathbf{z}(0)$ and $\mathbf{z}(S_k)$---reflecting the binary decision to include that item in the sum or not. Figure \ref{fig:subset}a demonstrates the resonator network successfully finding the solution when $T = 21$ and $|S| = 6$.

In order to use a residue number system, we need to choose $M$ so that $M > \sum_{s \in S} s$. This means that we need $\lceil\text{log} \ M \rceil$ bits per unit. This is an improvement from previous work using resonator networks to solve factorization problems \citep{kleyko2022integer}, which requires floating point precision to perform semi-prime factorization.

To understand the scaling capacity of the residue number system, we evaluate the performance of the resonator network as the set size increases. We observe that the resonator network finds exact solutions to the subset sum problem for large sets, and that performance improves with higher vector dimension (Figure~\ref{fig:subset}b). Figure~\ref{fig:subset}c illustrates that the success probability after up to 10 trials matches what is expected from 10 independent runs of the 1-trial accuracy. This finding suggests that the resonator network constitutes a `Las Vegas' algorithm \citep{babai1979monte}, in which each run has a success probability $p$, $p$ is independent across runs, and so the algorithm requires $1/p$ iterations on average. Accuracy also depends on the integer range searched over, even for the same set size (Figure~\ref{fig:subset}d), perhaps because larger integer ranges reduce the probability of multiple subsets matching the target. 

Finally, we compare our subset sum algorithm to brute force search and an exponential-time algorithm that solves the decision problem. The average number of iterations required by the resonator network to find a solution is drastically less than the exponentially increasing cost of a brute force search (Figure~\ref{fig:subset}e) and also improves with higher dimension. We find that on a CPU, the resonator network has faster clock time than the exponential time algorithm for $|S| > 28$ (Figure~\ref{fig:subset}f, Methods~\ref{sec:subsetsumexps}); most of the compute time is spent on generating the vector representations, rather than the resonator network dynamics itself. More significantly, whereas the baseline algorithm required $\mathcal{O}(2^{|S|})$ memory to keep candidate subsets in memory, the resonator network only requires $\mathcal{O}(D\cdot|S|)$ memory, since it never needs to explicitly represent every subset. We emphasize that the CPU implementation of the resonator network is primarily intended as a proof of concept, and that further performance gains would likely result from implementing the resonator network on emerging computing platforms, as in \citep{renner2022neuromorphic,langenegger2023memory}.

\section{Discussion}
\label{sec:discussion}

Our study provides the first definition of residue number systems with hyperdimensional computing. The framework inherits the benefits of both systems: the carry-free arithmetic and Chinese Remainder theoretic guarantees from residue number systems, along with the robustness and computing-in-superposition properties of hyperdimensional computing~\citep{KleykoComputingParadigm2022}. The framework provides a favorable way to encode, transform, and decode variables that is robust to noise. Taken together, these properties make Residue Hyperdimensional Computing an appealing framework for efficiently solving difficult computational problems, especially those involving combinatorial optimization. It also has implications for modeling population codes in the brain, in particular grid cells.

Prior work in computational neuroscience \citep{fiete2008grid,sreenivasan2011grid} has emphasized that residue number systems endow grid cells with useful computational properties, including high spatial resolution, modular updates, and error correction. We demonstrate that Residue Hyperdimensional Computing successfully achieves each of these coding properties (Sections~\ref{sec:residue_def}~\&~\ref{sec:scaling}) in a working neural implementation. Reciprocally, our algebraic framework makes two contributions to computational neuroscience. First, we show how to extend a residue number system to a self-consistent, non-negative hexagonal coordinate system. Second, we provide a new algorithm for collectively coupling spatial position to grid cell modules via the resonator network. The core prediction our framework makes for systems neuroscience is that each grid cell module corresponds to a factor estimate in the resonator network. More specifically, each module implements a toroidal attractor network, and multiplicative couplings with hippocampus and other grid cell modules enable error correction. This prediction is consistent both with recent experimental analysis supporting the existence of continuous attractor networks in single grid cell modules \citep{gardner2022toroidal}, and with recent theories of joint attractor dynamics in hippocampus and medial entorhinal cortex \citep{agmon2020theory}.

The framework also has implications for how neural populations can solve difficult optimization problems, such as disentangling of visual scenes. Recent work has emphasized the promise of hyperdimensional computing as an abstraction for neuromorphic computing \citep{frady2019robust,renner2022neuromorphic,KleykoComputingParadigm2022}. Residue Hyperdimensional Computing substantially reduces the storage and average number of operations needed for solving decoding problems and combinatorial optimization, contributing a simple yet powerful improvement. In addition, the phasor representations suggested by our framework directly map onto $Q$-state phasor networks \citep{noest1988discrete}, suggesting promising implementations in spiking neural networks \citep{bybee2022optimal} and strategies for solving combinatorial optimization problems \citep{wang2019oim}.

Finally, the performance of our framework on the subset sum problem suggests a new route for solving optimization problems with distributed representations and unconventional hardware. The subset sum problem is a particularly good fit for our framework because it is easily implemented by the Hadamard product operation on high-dimensional vectors.  Since other hard problems, such as 3-SAT, can be efficiently mapped to subset sum, our results potentially point the way to a new class of parallel algorithms for efficiently solving NP-hard problems.

\section{Methods}
\subsection{Definitions of algebraic operations}

\subsubsection{Definition of additive binding operation}
\label{sec:additivebinding}
We implement addition within Residue Hyperdimensional Computing by using the Hadamard product operation (component-wise multiplication, $\odot$) : i.e., $\mathbf{z}(x_1 + x_2) = \mathbf{z}(x_1) \odot \mathbf{z}(x_2)$. The Hadamard product correctly implements addition because it is commutative.  This can be seen as follows:
\begin{align*}
\mathbf{z}(x_1)\odot \mathbf{z}(x_2) &= (\mathbf{z}_{m_1}(x_1) \odot \mathbf{z}_{m_2}(x_1) \odot \dots \odot \mathbf{z}_{m_K}(x_1)) \odot (\mathbf{z}_{m_1}(x_2) \odot \mathbf{z}_{m_2}(x_2) \odot \dots \odot \mathbf{z}_{m_K}(x_2)) \\
&= (\mathbf{z}_{m_1}(x_1) \odot \mathbf{z}_{m_1}(x_2)) \odot (\mathbf{z}_{m_2}(x_1) \odot \mathbf{z}_{m_2}(x_2)) \odot \dots \odot (\mathbf{z}_{m_K}(x_1) \odot \mathbf{z}_{m_K}(x_2)) \\
&= \mathbf{z}_{m_1}(x_1 + x_2) \odot \mathbf{z}_{m_2}(x_1 + x_2) \odot \dots \odot \mathbf{z}_{m_K}(x_1 + x_2) \\
&= \mathbf{z}(x_1 + x_2).
\end{align*}

\subsubsection{Definition of multiplicative binding operation}
\label{sec:secondbinding}

To implement a second binding operation ($\star$), such that $\mathbf{z}(x_1 \cdot x_2)$ = $\mathbf{z}(x_1) \star \mathbf{z}(x_2)$, every component of the vector $\mathbf{z}(x_1)$ must be multiplied by $x_2$. If we had the value of $x_2$ explicitly, then we could directly implement $\mathbf{z}(x_1 \cdot x_2)$ by component-wise exponentiation of $\mathbf{z}(x_1)$ by $x_2$. However, decoding incurs additional computational costs, and we show here that multiplications can be computed without this intermediate step.

We require a few simplifying assumptions to define our multiplication operation. First, we assume that we have access to the individual base vectors for each moduli (e.g., $\mathbf{z}_{m_1}(x_1)$). If we do not, then we can use the resonator network to recover them. The key observation is that if $x$ is an integer, then each component of $\mathbf{z}_{m_k}(x)$ is itself a $m_k$-th root of unity. More specifically, it equals $e^{i\frac{2\pi}{m_k}r_j}$, for some integer $r_j = \frac{m_k}{2\pi}\phi_j x \ (\text{mod} \ m_k)$.

Therefore, we define an operation, $f$, that can multiply two discrete phases when they are both drawn from the $m_k$-th roots of unity: $f\left(e^{i\frac{2\pi}{m_k}r},e^{i\frac{2\pi}{m_k}s}\right) = e^{i\frac{2\pi}{m_k}rs}$. When $f$ is applied to two vectors of the same dimension, the multiplication is applied component-wise. Supposing that $r = \frac{m_k}{2\pi}\phi x_1$, and $s = \frac{m_k}{2\pi}\phi x_2$, we obtain $rs = \frac{m_k}{2\pi}\phi^2 x_1 x_2$, which is off from our desired result by a multiplicative factor of $\phi$. This motivates a final step of cancelling out this extra factor.

Because each phase $\phi$ is drawn from the $m_k$-th roots of unity, it can be written as $\frac{2\pi}{m_k}u$, where $u \in \mathbb{Z} \ ( \text{mod} \ m_k)$. When $m_k$ is prime, then any non-zero integer $u$ has a unique modular multiplicative inverse $v \in \mathbb{Z} \ ( \text{mod} \ m_k)$, such that $u \times v = 1$ (mod $m_k$). For example, the modular multiplicative inverse of 3 (mod 5) is 2. Consequently, $f\left(e^{i\frac{2\pi}{m_k}u^2},e^{i\frac{2\pi}{m_k}v}\right) = e^{i\frac{2\pi}{m_k}u}$. Motivated by these reasons, we therefore assume that whenever multiplicative binding is used, every moduli $m_k$ is prime. This assumption allows us to define an ``anti-base'' vector, $\mathbf{y}_{m_k}$, whose components are defined by the modular multiplicative inverses of $\mathbf{z}_{m_k}$. That is, if the $j$-th component of $\mathbf{z}_{m_k}$ is $e^{i\frac{2\pi}{m_k}u}$, then the $j$-th component of $\mathbf{y}_{m_k}$ is $e^{i\frac{2\pi}{m_k}v}$.

These assumptions motivate the following definition of the multiplicative operation, which we show successfully performs the multiplication of the arguments along with necessary cancellations:
\begin{align*}
    \mathbf{z}(x_1) \star \mathbf{z}(x_2) &:= f(f(\mathbf{z}_{m_1}(x_1),\mathbf{z}_{m_1}(x_2)),\mathbf{y}_{m_1}) \odot \dots \odot f(f(\mathbf{z}_{m_K}(x_1),\mathbf{z}_{m_K}(x_2)),\mathbf{y}_{m_K}) \\
    &= [e^{i \phi_{1,1}^2 x_1 x_2 \phi_{1,1}^{-1}},\dots,e^{i \phi_{1,D}^2 x_1 x_2 \phi_{1,D}^{-1}}] \odot \dots \odot [e^{i \phi_{K,1}^2 x_1 x_2 \phi_{K,1}^{-1}},\dots,e^{i \phi_{K,D}^2 x_1 x_2 \phi_{K,D}^{-1}}] \\
    &= [e^{i \phi_{1,1} x_1 x_2},\dots,e^{i \phi_{1,D} x_1 x_2}] \odot \dots \odot [e^{i \phi_{K,1} x_1 x_2},\dots,e^{i \phi_{K,D} x_1 x_2}] \\
    &= \mathbf{z}(x_1 \times x_2)
\end{align*}

We implement $f$ by extracting $s$ by taking the angle of the phasor $e^{i\frac{2\pi}{m_k}s}$, multiplying the angle by $\frac{m_k}{2\pi}$ and exponentiating $e^{i\frac{2\pi}{m_k}r}$ by the result. We compute modular multiplicative inverses via the built-in \textit{pow} function in Python. However, we note that both functions can also be implemented by lookup tables, and precomputing all input-output pairs may be optimal when many computations are re-used and lookups are inexpensive.

\subsection{Decoding methods}
In the context of high-dimensional distributed representations, the decoding problem is to recover a variable $x$ from a distributed representation $\mathbf{z}(x)$. In all of our decoding experiments, $x$ is either an integer or rational number. A survey of decoding methods, applied to symbolic hyperdimensional computing models, can be found in \cite{kleyko2023efficient}.

\subsubsection{Codebook decoding}

Codebook decoding estimates $x$ by taking inner products between $\mathbf{z}(x)$ and a precomputed set of reference vectors: $\hat{x} = \underset{x_k}{\operatorname*{arg\,max}} \langle \mathbf{z}(x),\mathbf{z}(x_k) \rangle $.

\subsubsection{Resonator network details}

The resonator network \citep{FradyResonator2020,kent2020resonator} is an algorithm for factoring an input vector, $\mathbf{z}$, into the primitives $\{ \mathbf{z}_1,\mathbf{z}_2,\dots,\mathbf{z}_K \}$ that compose it via Hadamard product binding: $\mathbf{z}=\mathbf{z}_1\odot\mathbf{z}_2\odot\dots\odot\mathbf{z}_K$. Each $\mathbf{z}_j$ is specified to come from a set of candidate vectors concatenated in a codebook, $\mathbf{Z}_j$, and therefore the search space grows combinatorially in terms of the product of codebook sizes.

The resonator network functions as a dynamical system with the following update equations:
\begin{equation}
    \hat{\mathbf{z}}_j(t+1) = g(\mathbf{Z}_{j}\mathbf{Z}_j^{\dagger} (\mathbf{z} \odot \prod_{i \neq j} \mathbf{z}_i^{\dagger}(t) ))
\end{equation}
\noindent
where $g$ is a non-linearity preserving phase and discarding angles of each complex component. In the case where we are decoding the representation of a residue number, the codebook for each factor (or moduli) $\mathbf{Z_j}$ is composed of $m_j$ entries, which are the vector encodings for the residues $[\mathbf{z}_{m_j}(0), \mathbf{z}_{m_j}(1), \dots, \mathbf{z}_{m_j}(m_j - 1)]$. In all experiments, we use an asynchronous update rule in which at each time step, only one factor estimate is updated, and every set of time steps, each vector is updated once. The algorithm runs either until convergence or until a maximum number of iterations has been reached. We consider the resonator converged when the normalized cosine similarity between two successive states exceeds a threshold, $\alpha$ (for all experiments, $\alpha = 0.95$).

\subsubsection{Evaluation of resonator network decoding accuracy, capacity and robustness to noise}

We evaluate resonator network accuracy as a function of vector dimension ($D$), effective range ($M$), number of moduli ($K$), and noise level (dependent on $\kappa$). We add noise only in experiments shown in Figure~\ref{fig:capacity}f, and $K = 2$ unless stated otherwise. $D=1024$ in Figure~\ref{fig:capacity}d, and $D=512$ in Figure~\ref{fig:capacity}f. To compute data points for curves that are a function of $M$, we generate a list of ascending primes, and select $K$ consecutive primes as moduli. The effective range, $M$, is the product of these moduli. We continue experiments for a fixed $D$ and increasingly large $M$ until empirical accuracy falls below a given threshold (0.95 for Figures~\ref{fig:capacity}a and ~\ref{fig:capacity}c, and 0.05 otherwise). To report the required number of comparisons for Figure~\ref{fig:capacity}a, we normalize the average number of inner product iterations by the accuracy, and visualize curves only in the high-accuracy regime (above 95 percent).

\subsection{Hexagonal residue encodings}
\label{sec:hexeq}

To project a $2D$ vector $\mathbf{x}$ to a $3D$ hexagonal coordinate $\mathbf{y}$, we multiply it by a matrix $\Psi$:
\[
\Psi=
  \begin{bmatrix}
    -\sqrt{3}/2 & -1/2 \\
    \sqrt{3}/2 & -1/2 \\
    0 & 1
  \end{bmatrix}
\]
The resulting vector $\mathbf{y}=\Psi\,\mathbf{x}$ is encoded as a high-dimensional vector using the generalization to multiple dimensions specified in Equation~\ref{eq:znfpe}. As an additional constraint, we require that $\mathbf{z}([1,1,1]) = \mathbf{z}([0,0,0])$. This condition implements the self-cancellation property (i.e, that moving equally along the three equiangular directions cancels out). It also converts possible negative values arising from the projection step to an equivalent non-negative coordinate encoding. 
Somewhat fortuitously, this constraint is naturally enforced in RHC by ensuring that for each component of $\mathbf{z}$, the three phases corresponding to the three directions sum to 0 (mod $2\pi$). This is achieved by constraining the joint probability distribution over triplets of phases so that this requirement is met (Figures~\ref{fig:2d}a and~\ref{fig:2d}c).

\subsection{Decoding with sub-integer precision}

\subsubsection{Extension of encoding scheme to rational numbers}
\label{sec:rationalfpe}

For a rational number $q \in \mathbb{Q}$, we define $\mathbf{z}_m(q)$ based on Fractional Power Encoding, modulo $m$ as:

\begin{equation}
    \mathbf{z}_m(q) = [e^{i \phi_{1}q}, e^{i \phi_{2}q}, \dots, e^{i \phi_{D}q}]
\end{equation}

\noindent We then form our representation of the remainders (modulo $m_k$) via the same process described in Equation~\ref{eq:residuehdc}. If $q$ is an integer, then this procedure matches that of Definition 2.2.1. But in general, $\mathbf{z}_m(q) \neq (\mathbf{z}_m)^q$. This is significant because while we can still evaluate similarity via inner products and perform addition operations, multiplication operations are no longer well defined.

\subsubsection{Sub-integer decoding with the resonator network}
\label{sec:subintdecoding}
Sub-integer decoding with the resonator network proceeds in three steps. First, we let the resonator update its factor estimates until convergence to a fixed point. We emphasize that sub-integer encodings are also fixed points of the resonator, even when the codebooks of the resonator contain only integers (Section \ref{sec:subint}). Second, we find the nearest integer codebook for each moduli, and generate the nearest codebooks for the fractional values within range 1 of that decoded integer ($r$ in total). Third, we use codebook decoding over these vectors encoding fractional values to return our result.

\subsubsection{Evaluation of sub-integer decoding with noise}
We fix $D=512$, and let $\kappa = \{16.0,1.0\}$. We run the resonator network until a maximum number of iterations or convergence, and evaluate if both the nearest integer and nearest fractional state are correct. If so, we regard the solution as correct, reporting accuracy and bits per vector.

\subsubsection{Measuring bits per vector}
\label{sec:measureinfo}

To measure the total amount of information decoded, we account for the accuracy of decoding and the number of states distinguished.  
The amount of information decoded for a single number (denoted as $I_{\mathrm{num}}$) is calculated using the corresponding accuracy ($a$) and size ($P$) of the total search space as
\begin{equation}
\label{eq:MI:item}
\begin{split}
I_{\mathrm{num}}(a, P) = & a \log_2(P a ) +  (1-a) \log_2 \left( \frac{P}{P-1} (1-a) \right).
\end{split}
\end{equation}
\noindent
For a detailed derivation of this equation, please refer to Section 2.2.3 of \cite{frady2018theory}. According to this metric, the amount of decoded information is $0$ when the accuracy is at chance ($1/P$).

\subsection{Visual scene factorization experiments}
\label{sec:csc}
Convolutional sparse coding learns a dictionary of basis functions, $\{\phi_j(x,y)\}$, and infers a set of sparse latent representations, $\{ A_j(x,y) \}$, for each image, $I(x,y)$, by minimizing the following energy function, \textit{E}:
\begin{equation}
    E = \frac{1}{2}|| I - \sum_{j=1}^{n}\phi_j \ast A_j ||^2_2 + \lambda \sum_{j=1}^{n}||A_j||_1
\end{equation}
\noindent
where $\ast$ denotes convolution, and $\lambda$ is a hyperparameter weighting the tradeoff between reconstruction error vs.\ sparsity. We use the SPORCO implementation of convolutional sparse coding introduced by \cite{wohlberg2017sporco} to learn the $\{\phi_j(x,y)\}$ for an ensemble of MNIST digits, and to infer the sparse representation $A$ for each image $I$. 

A useful feature of convolutional sparse coding is its equivariance to 2D translation; that is, 2D translation in the image domain results in 2D translation of the sparse representations, $\{ A_j(x,y) \}$. 
We can thus convert the set of sparse feature maps $\{ A_j(x,y) \}$ to a high-dimensional vector as follows:
\begin{equation}
    \mathbf{s} = \sum_{j,x,y} \textbf{h}(x) \odot \textbf{v}(y) \odot \textbf{d}_j \cdot A_j(x,y)
\label{eq:image-encoding}
\end{equation}
Here $\textbf{h}(x)$ and $\textbf{v}(y)$ denote the RHC encodings of horizontal ($x$) and vertical ($y$) position. $\mathbf{d}_j$ is a random vector generated i.i.d.\ that represents the identity of each basis function $\phi_j$. By expectation, most values of each $A_{j}(x,y)$ will be zero, because the energy function for sparse coding penalizes non-zero coefficients. Thus, the scene vector, $\mathbf{s}$, can be seen as a sparse superposition of position encodings of features ($\mathbf{\phi_i}$) contained in the image.

Now, we can separately define the vector encoding of each object $i$ to be recognized as 
\begin{equation}
    \mathbf{O}^{(i)} = \sum_{j,x,y} \textbf{h}(x) \odot \textbf{v}(y) \odot \textbf{d}_j \cdot o^{(i)}_j(x,y)
\end{equation}
where $o^{(i)}(x,y)$ is the sparse representation of the image of object $i$ within a {\em canonical reference frame}.  If we were to place object $i$ at position $(x^\prime,y^\prime)$ within an image, the resulting scene vector computed according to equation~\ref{eq:image-encoding} will be given as
\begin{equation}
\mathbf{s}=\mathbf{h}(x^\prime) \odot \mathbf{v}(y^\prime) \odot \mathbf{O}^{(i)}
\end{equation}
Therefore, our scene analysis problem amounts to one of factorizing $\mathbf{s}$ into its constituent vectors $\mathbf{h}(x^\prime)$, $\mathbf{v}(y^\prime)$, and $\mathbf{O}^{(i)}$.

We can factorize $\mathbf{s}$ using a resonator network with three codebooks, $\mathbf{O}, \mathbf{H},$ and $\mathbf{V}$.  Each element $\mathbf{O}^{(i)} \in \mathbf{O}$ consists of an encoding of each object as above, and $\mathbf{H}$ and $\mathbf{V}$ contain RHC encodings of horizontal and vertical position. 

For our object examples, we use 10 images from the MNIST dataset. Sparse coding dictionary elements are optimized over a subset of the MNIST dataset. After inferring a sparse code for each image, we encode it as a high-dimensional vector ($D=10,000$). We use a residue number system with bases $\{3,5,7\}$ for both horizontal and vertical dimension and then either enumerate all 105 codebooks for a single factor (Standard) or use 3 factors with 3, 5, and 7 codebooks, respectively (Residue). In either case, we ran the resonator network until convergence to a vector matching the scene representation (including reinitialization, if it did not converge after a fixed number of iterations or became stuck in a local minima), and record the average number of iterations multiplied by the average number of codebook evaluations (which is smaller for the residue encoding).

\subsection{Subset sum experiments}
\label{sec:subsetsumexps}
We use a residue number system with 3 moduli, $\{m-1,m,m+1\}$, where $m$ is a positive integer, ensuring that our moduli are co-prime. To generate random subset sum problems, we first define a maximum sum range to be $M/2$. For Figures~\ref{fig:subset}b and ~\ref{fig:subset}c, $m=200$, $M\approx 200^3$. Then, we draw random variables from a uniform distribution (scaled between 0, and half of the maximum sum over the largest set size tested). We then select a random subset of the set (all subsets are equally likely) and compute the sum. This sum forms the input to the resonator network, and we treat its solution is correct if it converged to the same sum. If the resonator network returns the wrong output, we restart it from a different random initialization, up to a maximum number of trials. We vary both the vector dimension ($D$) and set size ($|S|$), reporting accuracy after multiple simulations. For Figure~\ref{fig:subset}d, $D=400$.

To compare the number of evaluations relative to brute force (Figure~\ref{fig:subset}e), we record the average number of evaluations on each set size. We divide the number of inner product comparisons required for brute force evaluation by the number of comparisons per resonator network iteration. Further, we normalize the number of resonator iterations by the accuracy to ensure a fair comparison. In comparing our algorithm to a solver, we implement an exact subset-sum algorithm as a baseline \citep{nanda2005}. We let $m=1{,}000$, $D=\{10{,}000, 20{,}000\}$, and draw integers uniformly from the range $[0,5000]$.

\newpage
\printbibliography

\section*{Acknowledgments}
We thank Anthony Thomas, Eric Weiss, Josh Cynamon, and Amir Khosrowshahi for insightful discussions and feedback. The work of CJK was supported by the Department of Defense (DoD) through the National Defense Science \& Engineering Graduate (NDSEG) Fellowship Program. The work of DK, CB, FTS, and BAO was supported in part by Intel’s THWAI program. The work of CJK, CB, PK, and BAO was supported by the Center for the Co-Design of Cognitive Systems (CoCoSys), one of seven centers in JUMP 2.0, a Semiconductor Research Corporation (SRC) program sponsored by DARPA. DK has received funding from the European Union’s Horizon 2020 research and innovation programme under the Marie Sklodowska-Curie grant agreement No 839179. The work of FTS was supported in part by NIH under Grant R01-EB026955, and in part by NSF under Grant IIS-118991.

\section*{Corresponding authors}
Correspondence to \href{mailto:cjkymn@berkeley.edu}{Christopher J. Kymn} or \href{mailto:baolshausen@berkeley.edu}{Bruno A. Olshausen}.

\newpage

\setcounter{equation}{0}
\setcounter{figure}{0}
\setcounter{table}{0}
\setcounter{section}{0}
\setcounter{page}{1}
\makeatletter
\renewcommand{\thesection}{S-\Roman{section}}
\renewcommand{\theequation}{S.\arabic{equation}}
\renewcommand{\thefigure}{S.\arabic{figure}}
\renewcommand{\thetable}{S.\arabic{table}}
\resetlinenumber
\begin{appendix}
\begin{center}
    {\Huge\bf Supplemental material}
\end{center}

\section{A brief survey of distributed coding schemes}
\label{sec:otherlpe}

In order to process vector representations of numbers, such as in machine learning settings \citep{RachkovskijClassifiers2007, Rasanen2015tr,KleykoIndustrial2018,RahimiBiosignal2019,schindler2021primer,
kleykosurveyvsa2021part2}, previous work combined hyperdimensional computing with different kinds of locality-preserving encodings for representing numeric data with vectors. The requirement to be locality-preserving is that inner products between vectors encode similarity of the underlying data. Here we briefly review some locality-preserving encoding schemes that have been used in the past (see also \citep{kleykosurveyvsa2021part1}), assessing their kernel properties.

\begin{table}[h]
\resizebox{\textwidth}{!}{
\begin{tabular}{|l|c|c|c|c|c|}
\hline
Encoding scheme:                      & Algebra & Expressivity    & \makecell{Efficient \\ decoding} & \makecell{Robust \\ to noise} \\ \hline
One-hot                            & {\color{red} \xmark}                   & {\color{red} \xmark}       & {\color{cadmiumgreen} \checkmark}         &  {\color{red} \xmark}            \\ \hline
Thermometer \citep{PenzCloseness1987}, Supplement~\ref{sup:thermometer}                        & {\color{red} \xmark}                   & {\color{red} \xmark}       & {\color{cadmiumgreen} \checkmark}             & $\sim$            \\ \hline
\makecell[l]{Float \citep{GoltsevFloat1996}, Supplement~\ref{sup:float} \\ Gaussian Population Codes \citep{pouget2000information}} & {\color{red} \xmark}                   & {\color{red} \xmark}       & {\color{cadmiumgreen} \checkmark}             & $\sim$            \\ \hline
Scatter \citep{SmithScatter1990}, Supplement~\ref{sup:scatter}                            & {\color{red} \xmark}                   & {\color{red} \xmark}       & {\color{red} \xmark}               & {\color{cadmiumgreen} \checkmark}          \\ \hline
Fractional Power Encoding \citep{PlateNested1994}, Section \ref{sec:prelims}           & $\sim$             & {\color{cadmiumgreen} \checkmark} & {\color{red} \xmark}               &   {\color{cadmiumgreen} \checkmark}       \\ \hline \hline
\textbf{Residue Hyperdimensional Computing}               & {\color{cadmiumgreen} \checkmark}             & {\color{cadmiumgreen} \checkmark} & {\color{cadmiumgreen} \checkmark}         & {\color{cadmiumgreen} \checkmark} \\ \hline  
\end{tabular}
}
\caption{Existing high-dimensional vector-based schemes for encoding numbers (first five rows) in comparison to our proposed framework (last row).}
\label{table:compare}
\end{table}

\subsection{The thermometer code}
\label{sup:thermometer}

\begin{figure}[H]
\centering
\includegraphics[width=0.8\columnwidth]{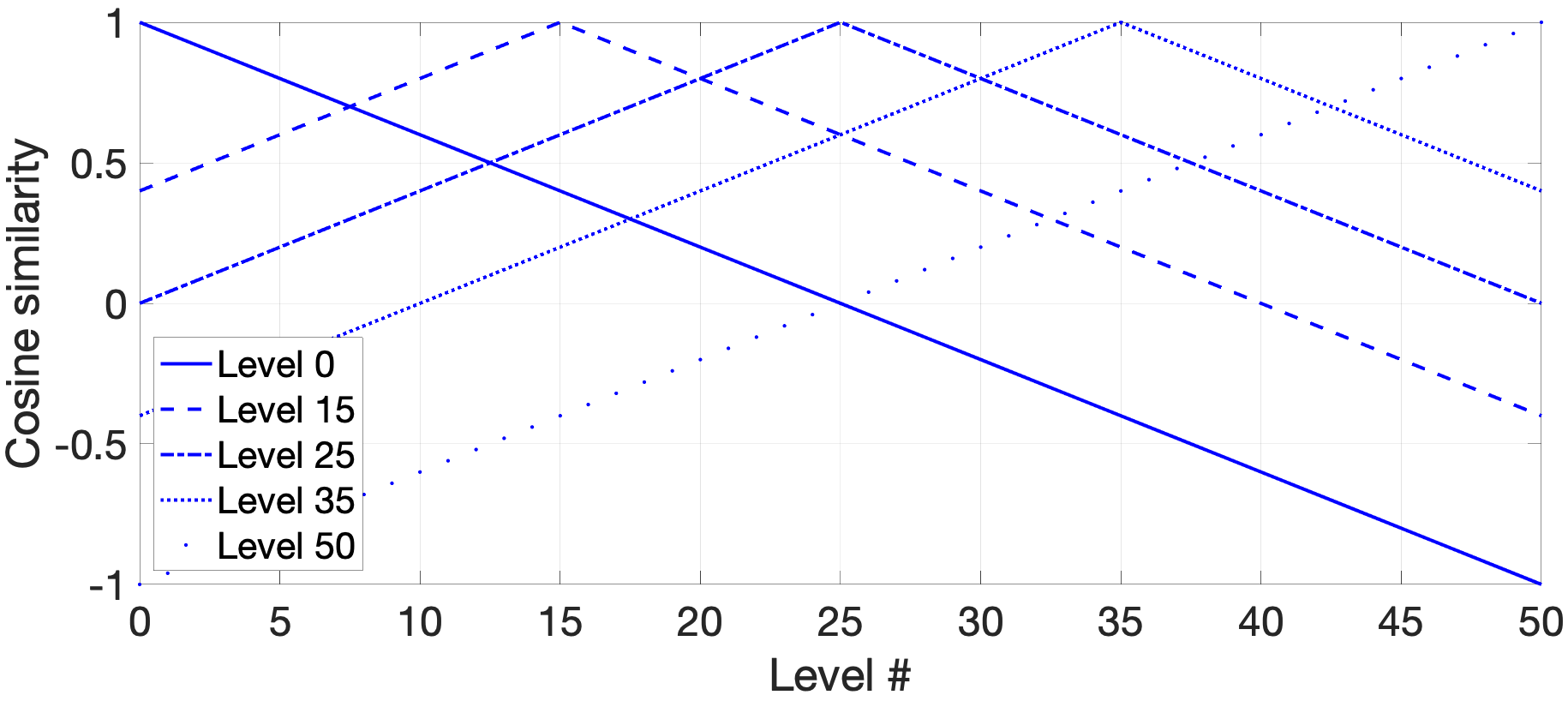}
\caption{
Similarity kernel of the thermometer code shown for several levels; $D$ was set to $50$.
}
\label{fig:thermometer}
\end{figure}

The thermometer code~\citep{PenzCloseness1987, RachkovskijScalars2005, BuckmanThermometer2018, KleykoDensityEncoding2020} is a simple and structured way to form a locality-preserving encoding for a range of discrete levels $s$, $s \in [0,D]$. 
The first code $\mathbf{z}(0)$ consists of all -1s. 
For other levels, the components of $\mathbf{z}(s)$ are determined as:
\begin{equation}
    z_i(s)= 
\begin{cases}
    +1,& i\leq s\\
    -1,              & \text{otherwise}
\end{cases}
\end{equation}
Thus, the last code $\mathbf{z}(D)$ consists of all +1s, and, in total, the thermometer code can represent $D+1$ levels. 
Figure~\ref{fig:thermometer} shows how cosine similarity appears for several different levels when $D=50$. Thermometer codes produce a translation-invariant kernel that is triangular and has a width of $2D+1$ levels. It is a nonlocal kernel, in the sense that there are no two points in the encoding range that have a similarity of zero. In practice, thermometer codes are used commonly when applying hyperdimensional computing to classification problems.
\subsection{The float code}
\label{sup:float}

\begin{figure}[H]
\centering
\includegraphics[width=0.8\columnwidth]{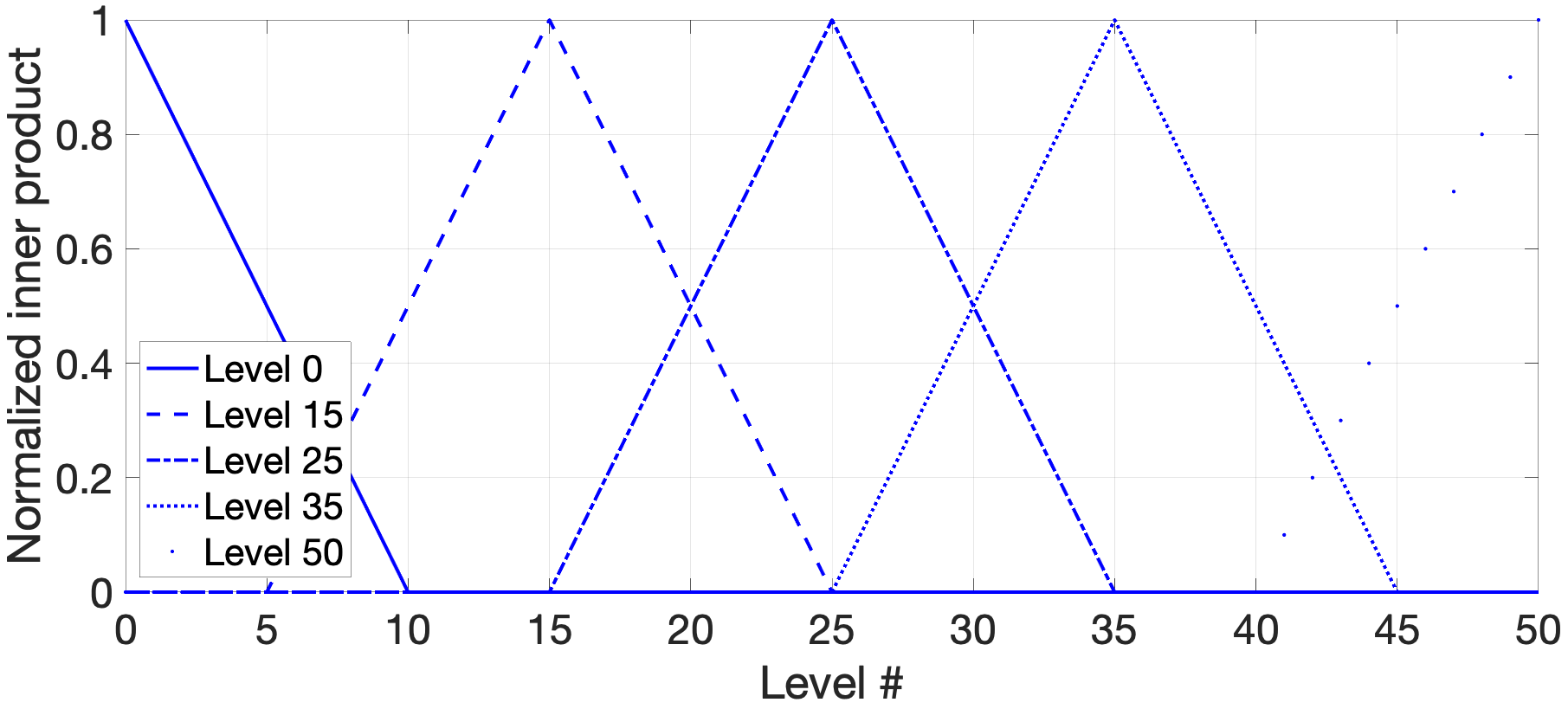}
\caption{
Similarity kernel of the float code shown for several levels; $D$ was set to $60$ while $w$ was set to $10$.
}
\label{fig:float}
\end{figure}

The float code, also known as the sliding code,~\citep{GoltsevFloat1996, RachkovskijScalars2005} addresses the issue of the thermometer code, i.e., that the similarity decay is not local. 
This is done by using $w$ consecutive +1 components (``float'') where the size of $w$ regulates similarity characteristics of the code. 
For the binary case, the similarity kernel of the float code is the triangular kernel of width $2w+1$ levels. 
To encode the lowest value $\mathbf{z}(0)$, the first $w$ components of the vector are set to +1s while the rest of the components are 0s.
In general, the components of $\mathbf{z}(s)$ are determined as:
\noindent
\begin{equation}
    z_i(s)= 
\begin{cases}
    +1, & s\leq i<s+w\\
    0,              & \text{otherwise}
\end{cases}
\end{equation}

Figure~\ref{fig:float} depicts how similarity (inner product normalized by $w$) decays for several levels in the float code for $D=60$, $w=10$. The float code also produces a triangular kernel, but
in contrast to the thermometer code, it allows controlling the width of the triangular kernel. 
The number of levels it could encode is still limited and equals to $n-w+1$. 

\subsection{The scatter code}
\label{sup:scatter}

\begin{figure}[H]
\centering
\includegraphics[width=0.8\columnwidth]{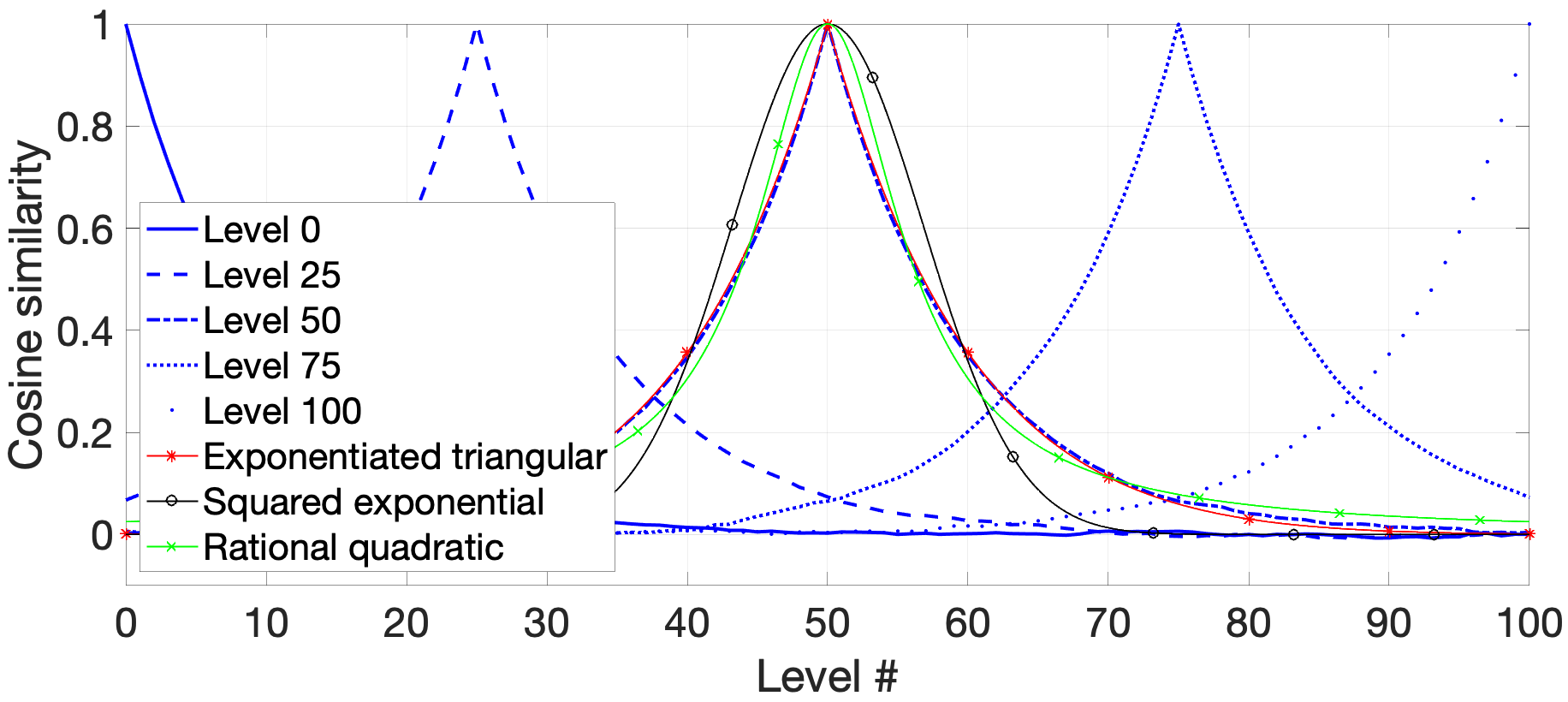}
\caption{
Similarity kernel of a scatter code; $D$ was set to $1000$, $p$ was $0.05$.
The values of similarities were averaged over $50$ random initializations of the code. 
}
\label{fig:scatter}
\end{figure}

Scatter codes~\citep{SmithScatter1990, RachkovskijScalars2005, kleyko2018classification} are another alternative to form a locality-preserving encoding where similarity decays nonlinearly.  
In scatter codes, the code for the first level $\mathbf{z}(0)$ is chosen randomly while each subsequent code is obtained from the previous one by randomly swapping its components with some probability $p$:
\begin{equation}
    z_i(s)= 
\begin{cases}
    -z_i(s-1),& r_i \leq  p  \\
    z_i(s-1),              & \text{otherwise}
\end{cases}
\end{equation}
where $r_i$ is a random value for the $i$-th component of $\mathbf{z}(s)$ chosen from the uniform distribution. 
Note that potentially there is no limitation on how many levels can be created with the scatter codes.

Figure~\ref{fig:scatter} shows how the cosine similarity looks for several different levels formed with the scatter code. Interestingly, the kernels are `bell-shaped'' with the exact shape depending on the parameter settings. 
To better figure out which standard kernel will correspond to this similarity, we have empirically fitted three kernels: exponentiated triangular: 
\begin{equation}
    K(s_1, s_2)= (1- \gamma |s_2-s_1|)^\alpha;
\end{equation}
squared exponential:
\begin{equation}
     K(s_1, s_2)= e^{-\frac{(s_2-s_1)^2}{2l^2}};
\end{equation}
and rational quadratic:
\begin{equation}
     K(s_1, s_2)= \left(1+\frac{(s_2-s_1)^2}{2\alpha l^2} \right)^{-\alpha} ;
\end{equation} 
The parameters of the kernels were chosen using the mean squared error as the fit criterion. 

\newpage
\section{Kernel properties of Residue Hyperdimensional \\ Computing}
To show that Fractional Power Encoding (modulo $m$) results in approximation of a particular periodic kernel with period $m$, we observe that our probability distribution can be written as a Dirac comb function pointwise multiplied by a $\text{rect}$ function. This fact becomes useful when we see that our kernel approximates a Fourier integral. Letting $x = x_1 - x_2$, we take the following steps to show convergence in the infinite-dimensional limit: 
\begin{align*}
    K_m^{*}(x_1,x_2) &= \lim_{D \to \infty} \frac{1}{D}\sum_{d=1}^D e^{i \phi_d x_1} \overline{e^{i \phi_d x_2}} \\
    &= \lim_{D \to \infty} \frac{1}{D} \sum_{d=1}^D e^{i \phi_d (x_1 - x_2)} \\
    &= \int e^{i \phi (x_1 - x_2)} p(\phi) d\phi \\
    &= \mathscr{F}^{-1} [ p(\phi) ] (x)\\
    &= \mathscr{F}^{-1} [\left(  \frac{1}{m} (\sum_{s \in \mathbb{Z}} \delta (\omega - \frac{2\pi}{m}s) \right) \cdot \left( \text{rect}(\frac{\omega}{2\pi}) \right)](x) \\
    &= [ \sum_{s \in \mathbb{Z}} \delta (x - ms) ] \circledast \text{sinc}(x) \\
    &= \sum_{s \in \mathbb{Z}} \text{sinc}(x - ms)
\end{align*}
\label{sec:periodic_kernel_math}

Thus, our kernel is a `sinc comb' function: a sum of sinc functions spaced with a period of $m$. This result is particularly notable because sinc evaluates to 0 for integers that are not a multiple of $m$ and means that distinct integers (and remainders) are orthogonal in the high-dimensional space.

To simplify the equation even further, we can derive a sum-less expression for an infinite number of sinc functions. We need to consider two cases: 1) $x = ms$ for some $s \in \mathbb{Z}$, and 2) $x \neq ms$ for all $s \in \mathbb{Z}$.

Case 1 is straightforward. Without loss of generality, let $x$ be the value for which $x-ms = 0$. Then we have:
\begin{align*}
    K_m(x) &= \sum_{s \in \mathbb{Z}} \text{sinc}(x - ms) \\
    &= 1 + \sum_{n \in \mathbb{N}} \text{sinc}(-mn) + \text{sinc}(mn) \\
    &= 1 \, \text{(since sinc evaluates to 0 for non-zero integers)}
\end{align*}

\noindent For case 2, without loss of generality, let $0 < |x| \leq m/2$. The answer differs subtly depending on whether $m$ is even or odd. Let us derive the even case first:
\begin{align*}
    K_m(x) &= \sum_{s \in \mathbb{Z}} \text{sinc}(x - ms) \\
    &= \sum_{s \in \mathbb{Z}} \frac{\text{sin}(\pi(x+ms))}{\pi(x+ms)} \\
    &= \frac{1}{\pi}\sum_{s \in \mathbb{Z}} \frac{\text{sin}(\pi x)}{x+ms} \\
    &= \frac{\text{sin}(\pi x)}{\pi}\sum_{s \in \mathbb{Z}} \frac{1}{x+ms} \\
    &= \frac{\text{sin}(\pi x)}{\pi m}\sum_{s \in \mathbb{Z}} \frac{1}{\frac{x}{m}+s} \\
    &= \frac{\text{sin}(\pi x)}{\pi m}\pi \text{cot}(\frac{\pi x}{m}) \\
    &= \frac{1}{m}\text{sin}(\pi x)\text{cot}(\frac{\pi x}{m})
\end{align*}
where the infinite sum is replaced due to an identity via the Herglotz trick: $\pi \text{cot}(\pi x) = \sum_{s \in \mathbf{Z}}\frac{1}{x+s}$. We can also use a second Herglotz identity, that $\pi \text{csc}(\pi x) = \sum_{n \in \mathbf{N}}\frac{(-1)^n}{x+n}$, to solve for the odd case:
\begin{align*}
    K_m(x) &= \sum_{s \in \mathbb{Z}} \text{sinc}(x - ms) \\
    &= \sum_{s \in \mathbb{Z}} \frac{\text{sin}(\pi(x-ms))}{\pi (x-ms)} \\
    &= \frac{\text{sin}(\pi x)}{\pi} \sum_{s \in \mathbb{Z}} \frac{(-1)^n}{x-ms} \\
    &= \frac{\text{sin}(\pi x)}{\pi} \sum_{s \in \mathbb{Z}} \frac{(-1)^n}{x+ms} \\
    &= \frac{\text{sin}(\pi x)}{\pi m} \sum_{s \in \mathbb{Z}} \frac{(-1)^n}{\frac{x}{m}+s} \\
    &= \frac{1}{m}\text{sin}(\pi x) \text{csc}(\frac{\pi x}{m})
\end{align*}

\begin{figure}[t]
\centering
\includegraphics[width=1.0\columnwidth]{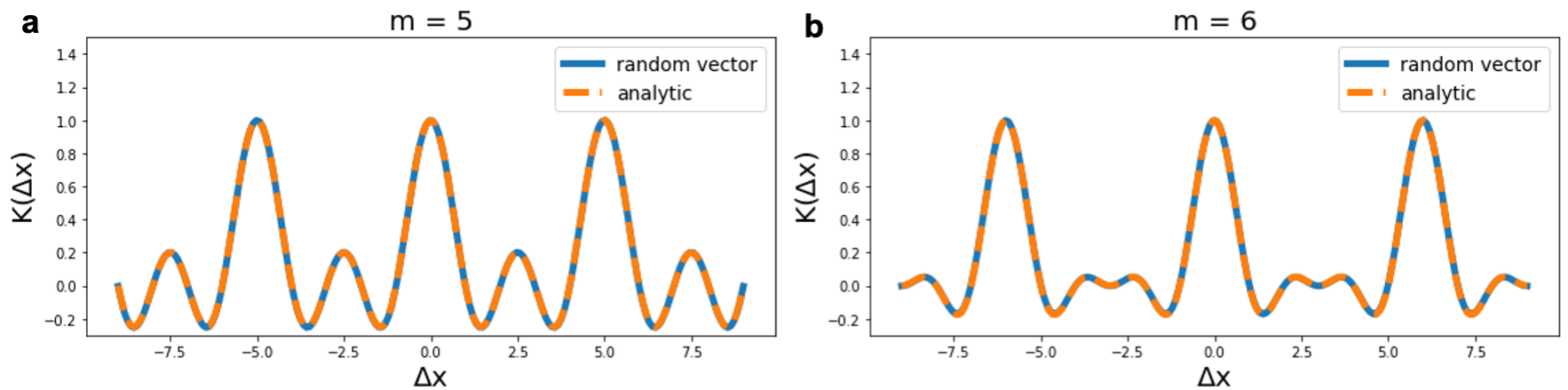}
\caption{\textbf{The analytic kernel expected by dashed lines matches the approximate kernel generated by a random vector of sufficiently high dimension} ($D$=50,000). \textbf{a}, match for an odd modulus ($m=5$), \textbf{b}, match for an even modulus ($m=6$).}
\label{fig:analyticsinc}
\end{figure}

\noindent We confirm our result by comparing our analytic values for both cases to the kernel induced by a high-dimensional vector (Figure~\ref{fig:analyticsinc}). In the limit of $m \to \infty$, both kernels converge to the sinc function: $K(x) = \frac{\text{sin}(\pi x)}{\pi x}$.

\end{appendix}
\end{document}